\documentclass{article} % For LaTeX2e
\usepackage[final]{colm2026_conference}
 
\usepackage{microtype}
\usepackage{hyperref}
\usepackage{url}
\usepackage{booktabs}
\usepackage{microtype}
\usepackage{graphicx}
\usepackage{subcaption}
\usepackage{booktabs} % for professional tables
\usepackage{algorithm}
\usepackage{algorithmic}
\usepackage{hyperref}
\usepackage{amsmath}
\usepackage{amssymb}
\usepackage{mathtools}
\usepackage{amsthm}
\usepackage{soul}
\usepackage{wrapfig}
\theoremstyle{plain}
\newtheorem{theorem}{Theorem}[section]

\theoremstyle{definition}

\theoremstyle{remark}
\newtheorem{remark}[theorem]{Remark}
 
\newlength{\reduce}
\usepackage{lineno}
 
\definecolor{darkblue}{rgb}{0, 0, 0.5}
\hypersetup{colorlinks=true, citecolor=darkblue, linkcolor=darkblue, urlcolor=darkblue}

\title{GRIP: Algorithm-Agnostic Machine Unlearning for Mixture-of-Experts via Geometric Router Constraints}
 
\author{Andy Zhu\textsuperscript{1}, Rongzhe Wei\textsuperscript{1}, Yupu Gu\textsuperscript{2}, Pan Li\textsuperscript{1}\\
\textsuperscript{1} Georgia Institute of Technology\\
\textsuperscript{2} Tsinghua University 
}
 
\begin{document}
 
\ifcolmsubmission
\linenumbers
\fi
 
\maketitle
 
\begin{abstract}
Machine unlearning in Mixture-of-Experts (MoE) large language models presents a critical yet underexplored challenge. Current unlearning methods applied to MoE architectures often exploit dynamic routing as an optimization shortcut: rather than genuinely erasing knowledge from expert parameters, they manipulate routers to redirect queries away from the originally assigned experts. This not only causes severe utility degradation but also leaves hazardous knowledge intact. Consequently, adversaries can bypass the router to recover sensitive information directly from dormant experts. In this study, we propose \textit{Geometric Routing Invariance Preservation (GRIP)}, an algorithm-agnostic framework that resolves these failure modes by enforcing hard geometric constraints on router updates. By projecting router gradient updates into the null space of the retain set's routing matrix, GRIP {suppresses} routing manipulation without freezing the router entirely, thereby {directing the unlearning pressure into the expert parameters themselves} across all relevant experts. GRIP offers two complementary variants: training-time stochastic projection and a post-training closed-form analytical correction. Extensive experiments on two MoE models across hazardous knowledge removal and copyright unlearning benchmarks demonstrate that GRIP restores routing stability from $0.21$ to $\geq 0.94$, improves retain accuracy by up to $89\%$, and reduces {white-box} adversarial knowledge recovery from $11\%$ to just {$3\%$ while in line with dense-architecture unlearning under black-box prompt attack,} establishing geometric constraints as a principled solution for genuine unlearning in sparse MoE architectures. \footnote{Code available at \url{https://github.com/AndyZhu311/GRIP}}

\end{abstract}

\color{black}
\section{Introduction}
\label{sec:intro}
 
Large language models (LLMs) have achieved remarkable progress across diverse domains~\citep{brown2020languagemodelsfewshotlearners, touvron2023llama2openfoundation, team2023gemini}, yet they can unintentionally memorize sensitive information from training data, including personal details, proprietary knowledge, and harmful content~\citep{carlini2021extracting, zhang2021counterfactual, nasr2023scalable,wei2024underestimated}. Machine unlearning (MU) has emerged as a principled solution, providing mechanisms to selectively remove such information from trained models without retraining from scratch~\citep{bourtoule2021machine, cao2015towards}, a capability essential for regulatory compliance with privacy laws such as GDPR's ``right to be forgotten''~\citep{gdpr2016, villaronga2018humans} and post-deployment safety~\citep{mitchell2022fast, weidinger2021ethical}. Existing efficient unlearning methods~\citep{yao2023editing, kurmanji2024towards, golatkar2020eternal, li2024wmdp, jia2024soul, korbak2023pretraining, zhang2024negative} typically optimize a forget objective (e.g., maximizing loss on target data) while preserving general model capabilities, and have demonstrated strong empirical performance on standard dense LLM architectures.
 
\begin{figure*}[t!]
  \centering
    \includegraphics[width = .9\textwidth]{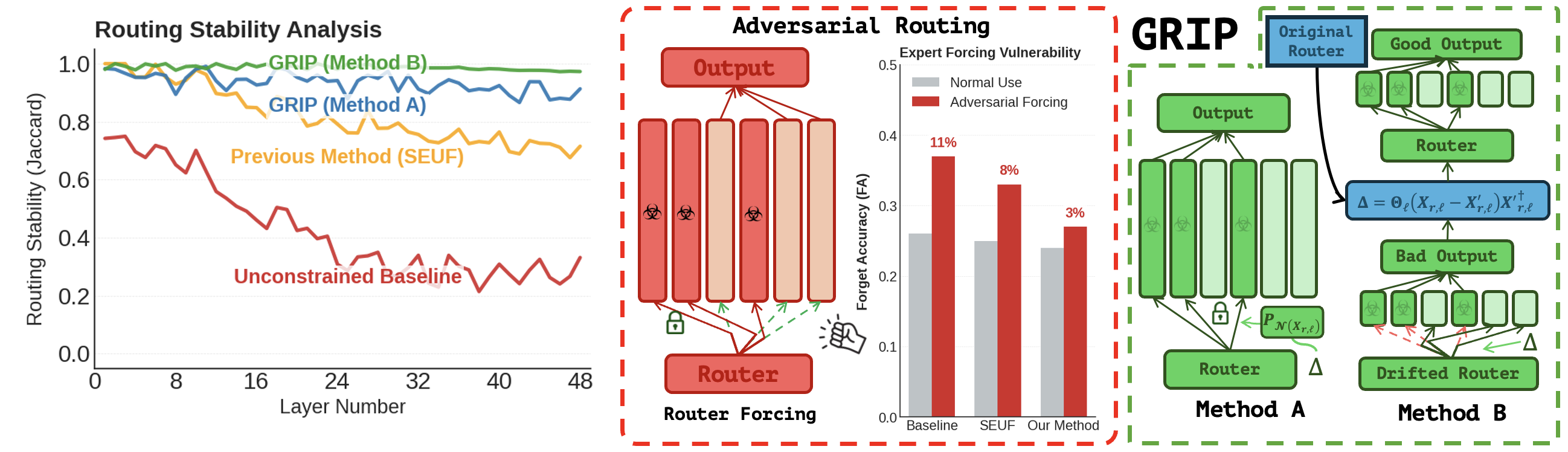}
  \caption{
\textbf{GRIP prevents expert selection shift and secures MoE models against 
adversarial knowledge recovery.} 
\textbf{(a) The Problem:} Standard unlearning exploits the router as a 
shortcut, diverting forget-set queries away from knowledgeable experts rather 
than erasing their parameters. Routing Stability (RS, Eq~\eqref{eq:jaccard}) 
collapses from $1.0$ to $\approx 0.2$ across layers.
\textbf{(b) Forcing Vulnerability:} Baselines merely suppress knowledge 
retrieval via routing; a white-box adversary can bypass this by directly 
forcing access to dormant experts, recovering sensitive information at 
a significantly higher rate than GRIP-unlearned models.
\textbf{(c) Our Solution:} GRIP imposes hard geometric constraints on router 
updates via Training-Time Null-Space Enforcement (\textit{Method A}) and 
Post-Training Analytical Realignment (\textit{Method B}), restoring routing 
stability ($\geq 0.94$) while achieving genuine knowledge erasure.
}
\label{fig:opening}
\end{figure*}

However, existing unlearning methods are primarily developed and validated on dense architectures, where all parameters engage uniformly with every input. This assumption breaks down in Mixture-of-Experts (MoE) models~\citep{jiang2024mixtral}, which are increasingly deployed in frontier systems due to their computational efficiency. In MoE layers, a learned router dynamically selects a sparse subset of expert networks for each input. Applying standard unlearning methods to these models can cause unstable expert selection for the knowledge that must be retained (see Fig.~\ref{fig:opening}(a)), severely degrading the model's utility. More critically, shifts in expert selection may leave the knowledge slated for unlearning intact within the original expert parameters. This introduces a potential risk: if adversaries can manipulate the router to access these dormant experts, they can easily recover the knowledge planned to be unlearned. We term this exploit an \textit{Expert Forcing} attack, a vulnerability unique to MoE models as, while suppressing access to knowledge without genuinely erasing it is a known problem in dense unlearning as well~\citep{erase_or_hide_2025, unlearning_reversibility_2025},  dense models lack unentangled pathways to route around for this paradigm. 
 
This vulnerability was first observed by SEUF~\citep{zhang2024seuf}, which attempts to address the resulting instability by applying soft regularization penalties and restricting unlearning updates to the single most relevant expert. However, empirical observations reveal that soft penalties offer limited protection against routing manipulation (see Fig.~\ref{fig:opening}(a)). Furthermore, updating only a single expert fails to eliminate hazardous knowledge stored across the other initially activated experts, leaving the model highly susceptible to the aforementioned Expert Forcing attacks.
 
To overcome these challenges, we propose \textit{Geometric Routing Invariance Preservation} (GRIP), a versatile framework for MoE unlearning that integrates with many existing optimization methods. GRIP imposes \textit{hard geometric constraints} on router updates to suppress expert selection shift and ensure unlearning pressure across all activated experts. Specifically, GRIP exerts fine-grained control over the routing mechanism by constraining router gradient updates to the null space defined by the retain set's relevant experts. This strict constraint substantially mitigates the routing instability responsible for utility collapse (Fig.~\ref{fig:opening}(a)). Concurrently, GRIP enables the optimization of both expert knowledge parameters and unselected routing paths. This broadens the optimization space compared to fixing all router paths, allowing the model to erase the targeted knowledge in a more thorough way. Consequently, GRIP achieves significantly stronger resistance to adversarial expert forcing attacks (Fig.~\ref{fig:opening}(b)). Algorithmically, GRIP offers two complementary realizations (Figure~\ref{fig:opening}(c)): an online approach (Method A) that enforces the constraint during training via stochastic projection, and an offline approach (Method B) that applies a single closed-form analytical correction after unlearning to realign drifted routers.

We validate GRIP on a 14.3B-parameter and a 30B-parameter MoE model (2.7B and 3B active parameters, respectively) across hazardous knowledge removal~\citep{li2024wmdp} and copyright unlearning~\citep{shi2024musemachineunlearningsixway}, instantiated with four distinct unlearning objectives. Across all settings, GRIP restores routing stability from $0.21$ - $0.45$ to at least $0.94$, improves retain accuracy by up to $89\%$, and reduces adversarial knowledge recovery under white-box expert forcing from $11\%$ to $3\%$, while remaining similarly resistant to black-box prompt attacks as dense-architecture unlearning. Together, these results show that hard geometric routing constraints provide a principled and scalable route to genuine unlearning in MoE models, where knowledge is erased from expert parameters rather than merely hidden by routing manipulation.

\section{Related Work}
\textbf{LLM Unlearning Methods.}
Machine unlearning for LLMs has been studied along several axes. A dominant family of methods operates through objective modification, including gradient ascent on forget data~\citep{jang2023knowledge}, gradient difference with retain regularization~\citep{kurmanji2024towards}, and negative preference optimization~\citep{zhang2024negative}. A second line of work targets internal representations directly, either by misdirecting activations away from harmful concepts~\citep{li2024wmdp} or by identifying and editing knowledge localized in specific layers or neurons~\citep{meng2022locating, wu2023depn}. More recent approaches explore parameter-efficient formulations via task arithmetic~\citep{ilharco2022editing} and second-order optimization~\citep{jia2024soul}, and in-context strategies that avoid weight modification entirely~\citep{pawelczyk2023context}. While these methods have demonstrated strong empirical performance, they are developed and validated exclusively on dense architectures, where all parameters engage uniformly with every input. This assumption is violated in sparse MoE models, motivating the need for unlearning methods specifically designed for dynamic routing architectures.
 
\textbf{Unlearning in Mixture-of-Experts Models.}
Sparse MoE architectures have become ubiquitous in frontier LLMs, powering systems such as Mixtral~\citep{jiang2024mixtral}, DeepSeek-V3~\citep{deepseekai2025deepseekv3technicalreport}, and GPT-4~\citep{openai2023gpt4}. Unlike dense models, MoE layers route each input through a learned sparse selection of experts, introducing a dynamic computation graph that poses fundamental challenges for post-training modification. To the best of our knowledge, the only prior work that targets MoE unlearning, SEUF~\citep{zhang2024seuf}, first identified that standard unlearning objectives tend to cause routers to shift expert selections, and proposed to mitigate this through soft regularization on router weights and to restrict parameter updates to the single highest-attributed expert. GRIP departs from it on two conceptual points rather than by tuning its penalty: we treat routing preservation as a feasibility condition satisfied by construction rather than as a soft penalty that discourages drift but admits it, and we treat erasure as inherently \textit{multi-expert}, since the targeted knowledge persists across the other initially-activated experts where it remains recoverable. These instantiate as a hard null-space projection in place of a soft $L_2$ penalty, and as leaving all expert weights free to update in place of a single-expert restriction. We also propose Expert Forcing, a diagnostic for whether knowledge survives in bypassed experts.
 
\textbf{MoE Routing Dynamics and Stability.}
Recent work has shown that MoE routing patterns encode meaningful semantic structure, with individual experts specializing in specific syntactic roles, domains, and reasoning patterns~\citep{gururangan2023scaling, zoph2022designing, fedus2022switch}, making routing preservation essential for maintaining model capabilities. While training-time stabilization techniques such as load balancing losses~\citep{fedus2022switch} and expert choice routing~\citep{zhou2022mixture} are well-established, they address a different setting than post-training unlearning. To our knowledge, GRIP is the first work to formalize routing {preservation} as a hard geometric constraint for machine unlearning. %, decoupling router preservation from parameter plasticity to enable genuine knowledge erasure in sparse architectures.

\section{Preliminaries}
\textbf{Machine Unlearning.} 
Given a model trained on an original dataset $D_{train}$, machine unlearning partitions this data into a forget set $D_f \subset D_{train}$ and a retain set $D_r = D_{train} \setminus D_f$. The objective is to produce updated model parameters that behave as if the model were trained exclusively on $D_r$. This retain-only oracle is unattainable for any scalable method, since retraining a frontier MoE from scratch is precisely the cost unlearning exists to avoid; Section~\ref{sec:conclusion} discusses what this implies for GRIP's objective.
 
\textbf{Mixture-of-Experts Layers.} A sparse MoE layer processes an input representation $x_\ell \in \mathbb{R}^d$ at layer $\ell$ using conditional computation, whereby only a fraction of the layer's parameters are activated per input. The layer contains $E$ expert networks and a router parameterized by $\Theta_\ell \in \mathbb{R}^{E \times d}$. For an input $x_\ell$, the router computes scores $s_\ell = \Theta_\ell x_\ell \in \mathbb{R}^E$ via linear projection and selects a subset of top-$k$ experts, $\mathcal{S}_\ell(x_\ell; \Theta_\ell) \subseteq \{1,\dots,E\}$. The layer output is: \begin{equation} \label{eq:moe_output} y_\ell = \sum_{j \in \mathcal{S}_\ell(x_\ell)} \frac{\exp(s_{\ell,j})}{\sum_{j' \in \mathcal{S}_\ell(x_\ell)} \exp(s_{\ell,j'})} \cdot f_j(x_\ell) \end{equation} where $f_j(x_\ell)$ denotes the output of the $j$-th selected expert, and the softmax normalization is computed only over the selected top-$k$ experts rather than all $E$ experts in the layer.
 
\textbf{Expert Selection Shift and Routing Stability.}
During unlearning, gradient-based methods update router parameters $\Theta_\ell \to \Theta_\ell + \Delta\Theta_\ell$, which can induce \textit{expert selection shift}: the set of selected experts changes from $\mathcal{S}_\ell^{\text{pre}}(x) := \mathcal{S}_\ell(x; \Theta_\ell)$ before unlearning to $\mathcal{S}_\ell^{\text{post}}(x) := \mathcal{S}_\ell(x; \Theta_\ell + \Delta\Theta_\ell)$ after, where we use $\mathcal{S}_\ell^{\text{pre}}$ and $\mathcal{S}_\ell^{\text{post}}$ as shorthands hereafter. As introduced in Section~\ref{sec:intro}, unlearning objectives can exploit this by manipulating $\Delta\Theta_\ell$ to redirect queries away from knowledgeable experts rather than erasing knowledge from their parameters. We formally quantify this effect using \textit{routing stability} (RS), the Jaccard similarity between pre- and post-unlearning expert selections over the retain set $D_r$: \begin{equation} \label{eq:jaccard} \text{RS}_\ell = \frac{1}{|\mathcal{Q}|} \sum_{x \in \mathcal{Q}} \frac{|\mathcal{S}_\ell^{\text{pre}}(x) \cap \mathcal{S}_\ell^{\text{post}}(x)|} {|\mathcal{S}_\ell^{\text{pre}}(x) \cup \mathcal{S}_\ell^{\text{post}}(x)|} \end{equation} where $\mathcal{Q} \subseteq D_r$ is the set of evaluation queries drawn from the retain set. An RS of 1 indicates perfect routing preservation, while 0 indicates complete routing collapse. Note that RS measures fidelity to the \textit{original} model's routing, not to a retain-only oracle's (Section~\ref{sec:conclusion}).

\section{Methodology}
\label{sec:method}
In standard dense LLMs, unlearning objectives operate over a static computation graph where all parameters engage uniformly with every input. MoE architectures fundamentally alter this landscape: the computation path through expert networks is itself a function of the router parameters $\{\Theta_\ell\}$, creating a dual optimization landscape where the unlearning objective $\mathcal{L}_{\text{MU}}$ can be minimized via two distinct strategies, either by (1) genuinely erasing knowledge from expert parameters $W$, or (2) manipulating the router to redirect forget-set queries away from knowledgeable experts entirely. Critically, the router is the cheaper descent direction, since a single linear map gates every expert in the layer, and we observe correspondingly larger norms for $\nabla_{\Theta}\mathcal{L}_{\text{MU}}$ than $\nabla_{W}\mathcal{L}_{\text{MU}}$, creating a structural bias toward the routing shortcut. Without proper constraints, the optimizer will therefore preferentially converge to strategy (2), producing the expert selection shift and superficial forgetting. 
 
GRIP addresses this by imposing a hard constraint directly on router updates. Formally, given any base unlearning objective $\mathcal{L}_{\text{MU}}(D_f, D_r)$, GRIP augments the optimization with a \textit{retain-side routing invariance constraint}: router parameter updates $\Delta\Theta_\ell$ must preserve the expert selections of the retain set $D_r$, ensuring that $\mathcal{S}_\ell^{\text{post}}(x) = \mathcal{S}_\ell^{\text{pre}}(x)$ for all $x \in D_r$. By making the routing shortcut infeasible on the constrained inputs, this pushes the optimization to achieve $\mathcal{L}_{\text{MU}}$ through knowledge erasure within expert weights instead. Because expert parameter updates remain unconstrained, unlearning can proceed across \textit{all} relevant experts rather than being restricted to a single expert as the existing work does~\citep{zhang2024seuf}. We instantiate this principle via two variants: \textit{Method A} enforces the constraint online at each training step via stochastic gradient projection (Sec.~\ref{sec:method_a}), while \textit{Method B} applies a single closed-form analytical correction after unlearning to realign drifted routers (Sec.~\ref{sec:method_b}). The constraint is placed on the retain side in order both to preserve model utility and, because router rows are shared across forget and retain inputs, to limit how freely the router can reroute forget queries through that same expert; Appendix~\ref{app:constraint_side} compares this against forget-side and combined alternatives.
 
% The derivation of the constraint itself is presented in Section~\ref{sec:relaxed}.
 
\subsection{From Routing Invariance to Selection-Preserving Constraints}
\label{sec:relaxed}
 
We need to avoid potentially too restrictive constraints on the router. The retain-side routing invariance constraint requires $\mathcal{S}_\ell^{\text{post}}(x) = \mathcal{S}_\ell^{\text{pre}}(x)$ for all $x \in D_r$. Since the top-$k$ selection function is piecewise constant and non-differentiable with respect to $\Delta\Theta_\ell$, we seek a differentiable sufficient condition. Let $X_{r,\ell} \in \mathbb{R}^{d \times N_r}$ denote the matrix of retain representations at layer $\ell$, where each column $x^{(i)}_{r,\ell} \in \mathbb{R}^d$ is the hidden state of the $i$-th retain input at that layer. A natural sufficient condition is logit invariance $\Delta\Theta_\ell X_{r,\ell} = 0$, requiring that any router update resides in $\text{Null}(X_{r,\ell}^\top)$. However, this global constraint is overly restrictive: routing stability requires only that the top-$k$ selection order is preserved, not that all logits remain static. Enforcing zero logit change for every retain input unnecessarily prohibits benign modifications to non-selected experts, severely limiting the feasible optimization landscape and inducing exactly the parameter rigidity that GRIP is designed to avoid.
 
\textbf{Finer-grained Constraints.} To resolve this, we decompose the global constraint into an asymmetric, expert-specific formulation. Instead of rigidly fixing the entire routing distribution, we strictly preserve only the top-$k$ selections. The routing paths for non-selected experts can be freely adjusted, provided their scores do not exceed the threshold to become a top-$k$ expert. For expert $j$ at layer $\ell$, let $\mathcal{I}_{j,\ell} = \{i : j \in \mathcal{S}_\ell^{\text{pre}}(x^{(i)}_{r,\ell})\}$ denote the retain inputs for which expert $j$ was selected. We impose: 
\begin{equation} \label{eq:relaxed_constraint} (\Delta\Theta_\ell)_{j,:}\, x^{(i)}_{r,\ell} = 0 \;\; \forall i \in \mathcal{I}_{j,\ell}, \qquad (\Delta\Theta_\ell)_{j,:}\, x^{(i)}_{r,\ell} \leq \tau_{i,j,\ell} \;\; \forall i \notin \mathcal{I}_{j,\ell} 
\end{equation} 
where $s_k = (\Theta_\ell)_{k,:}\, x^{(i)}_{r,\ell} \in \mathbb{R}$ denotes the routing score of expert $k$ for input $x^{(i)}_{r,\ell}$ under the pre-unlearning router, and $\tau_{i,j,\ell} = \min_{k \in \mathcal{S}_\ell^{\text{pre}}(x^{(i)}_{r,\ell})} [s_k - s_j]$ is the selection margin of expert $j$ on input $i$, i.e., the minimum score gap between any selected expert and expert $j$ itself. The equality constraint preserves routing decisions for inputs that selected expert $j$, while the inequality constraint permits score modifications for non-selected inputs, so long as they do not breach the selection margin. Together, these conditions ensure that $\mathcal{S}_\ell^{\text{post}}(x) = \mathcal{S}_\ell^{\text{pre}}(x)$ on every $x \in D_r$ over which they are actually enforced, while substantially relaxing the constraints for genuine knowledge erasure. 
 
However, enforcing these constraints exactly over the full retain set $D_r$ requires projecting onto a global polytope satisfying $\mathcal{O}(|D_r| \times E)$ simultaneous constraints while caching $\mathcal{O}(|D_r| \times d)$ intermediate representations, which is strictly intractable at LLM scale. The preservation we obtain is therefore empirical rather than worst-case, which motivates the batch-level approximations developed in the following sections.
 
\subsection{Training-Time Enforcement via Stochastic Projection}
\label{sec:method_a}
 
Equation~\eqref{eq:relaxed_constraint} gives the ideal selection-preserving constraints for a router update. Enforcing these constraints exactly over the full retain set at every optimization step is intractable, so a training-time variant applies them at the level of the current retain minibatch. For simplicity, in this subsection we reuse $X_{r,\ell}$ and $\mathcal{I}_{j,\ell}$ to denote the retain representations and selected-example index set for the current minibatch at layer $\ell$.
 
For expert $j$ at layer $\ell$, let $u_{j,\ell} \in \mathbb{R}^d$ denote the unconstrained update proposed for the $j$-th router row by the base unlearning algorithm. GRIP replaces this update by a projected update $\bar{u}_{j,\ell}$ that lies in the minibatch feasible set:
\begin{equation}
\label{eq:batch_feasible}
\mathcal{K}_{j,\ell} = \Big\{ u \in \mathbb{R}^d : \langle u, x^{(i)}_{r,\ell} \rangle = 0 \;\; \forall i \in \mathcal{I}_{j,\ell}, \; \langle u, x^{(i)}_{r,\ell} \rangle \le \tau_{i,j,\ell} - \varepsilon \;\; \forall i \notin \mathcal{I}_{j,\ell} \Big\},
\end{equation}
where $\varepsilon > 0$ is a safety margin that improves numerical robustness and prevents tie-induced changes in the top-$k$ selected set. We enforce membership in $\mathcal{K}_{j,\ell}$ in two stages.
 
\paragraph{Stage 1: Equality projection.}
We first enforce the hard equality constraints for the inputs that originally selected expert $j$. Let $X^{\mathrm{eq}}_{j,\ell} = [x^{(i)}_{r,\ell}]_{i \in \mathcal{I}_{j,\ell}} \in \mathbb{R}^{d \times |\mathcal{I}_{j,\ell}|}$ collect the corresponding retain representations. The orthogonal projector onto the complement of their span is 
%\begin{equation}
%\label{eq:eq_projector}
$P^{\mathrm{eq}}_{j,\ell} = I - X^{\mathrm{eq}}_{j,\ell} \Bigl(X^{\mathrm{eq}\top}_{j,\ell} X^{\mathrm{eq}}_{j,\ell}\Bigr)^{\dagger} X^{\mathrm{eq}\top}_{j,\ell},$ 
%\end{equation}
and the intermediate update is 
%\begin{equation}
%\label{eq:eq_update}
$\tilde{u}_{j,\ell} = P^{\mathrm{eq}}_{j,\ell} u_{j,\ell}.$
%\end{equation}
By construction, $\langle \tilde{u}_{j,\ell}, x^{(i)}_{r,\ell} \rangle = 0$ for all $i \in \mathcal{I}_{j,\ell}$, so the score of expert $j$ is unchanged on the retain inputs that actively use it.
 
\paragraph{Stage 2: Inequality correction.}
The update $\tilde{u}_{j,\ell}$ must still satisfy the margin constraints for retain inputs that did not originally select expert $j$. For each $i \notin \mathcal{I}_{j,\ell}$, define the half-space $\mathcal{H}_{i,j,\ell} = \{ u \in \mathbb{R}^d : \langle u, x^{(i)}_{r,\ell} \rangle \le \tau_{i,j,\ell} - \varepsilon \}$. Exact Euclidean projection of $\tilde{u}_{j,\ell}$ onto the full intersection $\bigcap_{i \notin \mathcal{I}_{j,\ell}} \mathcal{H}_{i,j,\ell}$ remains expensive, so we approximate it with a stochastic alternating-projection procedure. If a sampled input $i \notin \mathcal{I}_{j,\ell}$ violates the margin, i.e., $\langle \tilde{u}_{j,\ell}, x^{(i)}_{r,\ell} \rangle > \tau_{i,j,\ell} - \varepsilon$, the Euclidean projection onto the boundary of $\mathcal{H}_{i,j,\ell}$ is 
%\begin{equation}
%\label{eq:rk_update}
$\tilde{u}_{j,\ell} \leftarrow \tilde{u}_{j,\ell} - \frac{\langle \tilde{u}_{j,\ell}, x^{(i)}_{r,\ell} \rangle - (\tau_{i,j,\ell} - \varepsilon)}{\|x^{(i)}_{r,\ell}\|_2^2} x^{(i)}_{r,\ell}.$
%\end{equation}
We implement this stage using a randomized Kaczmarz-style update for systems of linear \textit{inequalities} \citep{leventhal2010randomized, deloera2017sampling}, an efficient stochastic approximation to the projection onto the polyhedral feasible region. Further discussion of this algorithm is in Appendix~\ref{app:halfspace}.
 
After these stages, the resulting $\bar{u}_{j,\ell}$ is used as the update for the $j$-th router row. Importantly, GRIP modifies only router updates; expert weights $W$ and other parameters remain exactly those produced by the standard unlearning algorithms such as gradient difference or NPO. Imposing these hard constraints incurs a $1.71\times$ training overhead. Constraint storage requirements are negligible due to a tiered low-rank caching scheme that holds the active memory footprint below $1\%$ of model size; further analysis of computational and storage costs is contained within Appendix~\ref{app:memory}.

\subsection{Post-Training Router Realignment}
\label{sec:method_b}
 
The training-time variant enforces routing invariance at every step, but faces a limitation in deep multi-layer models: representation drift. As unlearning updates parameters in layers $1, \dots, \ell-1$, the retain representations $X_{r,\ell}$ shift to $\tilde{X}_{r,\ell}$, rendering precomputed null-space projectors increasingly stale in deeper layers. Rectifying this by continuously recomputing projectors necessitates $\mathcal{O}(Ed^3)$ operations per layer per update step, associating exact online enforcement with high computational cost. 
 
% Further, this drift exposes that simply freezing router weights does not freeze the routing decisions and is thus not a viable solution(Section~\ref{sec:ablation}).
 
The post-training variant addresses this by deferring enforcement entirely. Instead of projecting router updates at every training step, we first run the chosen unlearning algorithm without router constraints, then apply a single post-hoc correction that restores the router's behavior on the retain set. Let $\Theta_\ell$ and $X_{r,\ell}$ denote the pretrained router and retain representations at layer $\ell$, and let $\tilde{\Theta}_\ell$ and $\tilde{X}_{r,\ell}$ denote their counterparts after unconstrained unlearning. We seek a correction $\Delta_\ell$ such that the original routing scores are restored on the post-unlearning retain representations: $(\tilde{\Theta}_\ell + \Delta_\ell)\tilde{X}_{r,\ell} = \Theta_\ell X_{r,\ell}$. This yields the closed-form least-squares solution:
\begin{equation}
\label{eq:correction}
\Delta_\ell = \Theta_\ell X_{r,\ell}\tilde{X}_{r,\ell}^{\dagger} - \tilde{\Theta}_\ell, \qquad \tilde{X}_{r,\ell}^{\dagger} = \tilde{X}_{r,\ell}^{\top} \bigl( \tilde{X}_{r,\ell}\tilde{X}_{r,\ell}^{\top} + \lambda I \bigr)^{-1},
\end{equation}
where $\lambda = 10^{-6}$ is a small constant for numerical stability (further details in Appendix~\ref{app:complexity}).
 
The efficacy of this correction relies on the observation that unlearning induces bounded perturbations in representation space, such that the drift $X_{r,\ell} \to \tilde{X}_{r,\ell}$ preserves the local topology of the retain manifold. This assumption holds almost universally except for slight breakdowns under very aggressive forgetting, at deep layers, or for forget sets entangled with the retain set; the acceptance check of Appendix~\ref{app:complexity} flags the layers where it fails, which fall back to Method A. Consequently, the router's linear decision boundaries do not need to be re-learned from scratch but simply require geometric realignment to the shifted input distribution, which is exactly what Equation~\eqref{eq:correction} achieves. This variant replaces per-step projections by a single closed-form correction per layer, reducing the projection cost from $\mathcal{O}(KLEd^3)$ to $\mathcal{O}(Ld^3)$ and the total overhead from $1.71\times$ to $1.21\times$ training time; Appendix~\ref{app:complexity} gives the full cost and memory analysis.

\section{Experimental Results} 
\label{sec:results}
 
\textbf{Datasets and Metrics.} We adopt Qwen3-30B-A3B~\citep{qwen3technicalreport} and Qwen1.5-MoE-A2.7B-Chat~\citep{bai2023qwentechnicalreport} as the backbones. We evaluate on the WMDP~\citep{li2024wmdp} and MUSE~\citep{shi2024musemachineunlearningsixway} benchmarks using standard protocols as detailed by their original papers, where we approximately controlled the forget metrics (Appendix~\ref{app:eval_protocol}), i.e., all models were unlearned until they matched the baseline level for forget metrics, to provide the most accurate comparison on retain ability. We report Forget Accuracy (FA) and Retain Accuracy (RA) for the WMDP dataset and metrics C1 through C4 for the MUSE dataset as well as Routing Stability (RS) as defined in Eq.~\eqref{eq:jaccard} (see Appendix~\ref{app:eval_protocol} for more details on training and metrics).  %A detailed breakdown of these metrics is located in Appendix~\eqref{app:eval_protocol}.
\subsection{Main Results}
Table~\ref{tab:main_results} presents results across four unlearning objectives (GD, KL, NPO, RMU) on both WMDP-Cyber and MUSE benchmarks, where forget metrics are controlled to approximately matched levels to isolate differences in utility preservation. We organize our analysis around three findings: (i)~GRIP consistently restores routing stability to near-perfect levels, (ii)~this stability translates directly into superior retain accuracy that approaches dense-model baselines, and (iii)~GRIP subsumes SEUF across all settings.
 
\begin{table*}[t!]
\centering
\caption{Comprehensive evaluation across WMDP and MUSE benchmarks. $\downarrow$/$\uparrow$ indicates lower/higher is better. Best methods in each category are \textbf{bolded}. Best overall method is \hl{highlighted}. Statistics that were controlled are \underline{underlined}. Router stability values calculate Jaccard similarity in expert selections using Equation ~\eqref{eq:jaccard}.
}
\label{tab:main_results}
\small
\setlength{\tabcolsep}{3.5pt}
\begin{tabular}{lccc|ccccc}
\toprule
& \multicolumn{3}{c|}{\textbf{WMDP-Cyber}} & \multicolumn{5}{c}{\textbf{MUSE}}\\
\cmidrule{2-9}
\textbf{Method} & \underline{FA}$\downarrow$ & RA$\uparrow$& RS $\uparrow$ & \underline{C1}$\downarrow$ &\underline{C2}$\downarrow$ & \underline{C3}$\,{\to}\,0$& C4$\uparrow$ & RS $\uparrow$\\
\midrule
GD & 0.26 & 0.32 & 0.21 & 0.05 & 0.03 & -25.6 & 4.31 & 0.26\\
 {SEUF + GD} & 0.27 & 0.44 & 0.90 & 0.07 & 0.09 & -21.4 & 8.54 & 0.92\\
 
{Constrained + GD} &  \textbf{0.24} &  {0.51} &  {0.94} &  {0.04} &  {0.02} &  {-26.8} &  \textbf{11.4} &  {0.96 }\\
 {PTC + GD} &  {0.26} &  \textbf{0.55} &  \textbf{0.99} &  \textbf{0.01} &  \textbf{0.00} &  -30.4 &  {9.42} &  \textbf{1.00}\\
\midrule
KL & 0.28 & 0.35 & 0.42 & \textbf{15.3} & 22.9 & -33.5 & 21.3 & 0.32\\
 {SEUF + KL} & 0.25 & 0.49 & 0.89 & 18.5 & 30.2 & -32.4 & 37.6 & 0.91\\
 {Constrained + KL} &  {0.27} &  {0.58} &  {0.94} &  {17.2} &  {24.6} &  {-30.1} &  \hl{\textbf{42.7}} &  {0.96}\\
 {PTC + KL} &  \textbf{0.26} &  \textbf{0.59} &  \hl{\textbf{1.00}} &  {16.6} &  \textbf{22.4} &  {-32.9} &  {41.8} &  \textbf{1.00}\\
\midrule
NPO & 0.33 & 0.48 & 0.45 & 16.3 & \textbf{22.1} & -39.5 & 12.6 & 0.43\\
 {SEUF + NPO} & 0.34 & 0.53 & 0.91 & 14.2 & 28.7 & -35.3 & 35.1 & 0.89\\
 {Constrained + NPO} &  {0.29} &  \textbf{0.62} &  {0.97} &  \textbf{15.3} &  {23.8} &  {-42.3} &  \textbf{41.6} &  {0.94}\\
 {PTC + NPO} &  \textbf{0.25} &  {0.59} &  \textbf{0.99} &  {19.3} &  {27.3} &  -36.1 &  {41.2} &  \textbf{1.00}\\
\midrule
RMU & 0.26 & 0.35 & 0.43 & 0.00 & 0.01 & -41.3 & 14.2 & 0.37\\
 {SEUF + RMU} & 0.25 & 0.58 & 0.89 & 0.00 & 0.03 & -40.8 & 35.1 & 0.91\\
 {Constrained + RMU} &  \textbf{\hl{0.24}} &  {0.65} &  {0.96} &  {0.02} &  \hl{\textbf{0.00}} &  -42.8 &  \textbf{40.5} &  {0.95}\\
 {PTC + RMU} &  {0.28} &  \hl{\textbf{0.66}} &  \textbf{0.99} &  \hl{\textbf{0.00}} &  {0.02} &  {-41.9} &  {39.2} &  \hl{\textbf{1.00}}\\
\midrule
 {Dense Baseline} & 0.28 & 0.70 & N/A & 17.0 & 25.0 & -43.5 & 44.6 & N/A\\
\bottomrule
 
\end{tabular}
\end{table*}
 
\textbf{Significant Routing Stability Improvements.} GRIP maintains consistent near-perfect routing stability (RS $\geq$ 0.94) across all base unlearning methods. While baseline approaches suffer severe routing disruption, with stability dropping as low as 0.21, our post-training correction (PTC) consistently achieves \textbf{perfect or near-perfect stability (0.99-1.00)} and the online constraints also maintain high stability (0.94-0.97). %Because routing stability governs which experts process retain-set queries, these gains translate into utility improvements.
 
\textbf{Superior Knowledge Retention With Unlearning Effectiveness.} The above routing stability naturally translates into utility improvements. Our framework demonstrates substantial gains in retain accuracy, improving 23--89\% relative to baseline methods. For instance, on WMDP-Cyber with RMU, retain accuracy improves from 0.35 to 0.65--0.66 (+86--89\%). Critically, this routing preservation does not compromise unlearning quality, as all GRIP variants reach forget accuracy at or below the baseline, confirming that the utility gains stem from eliminating the routing shortcut rather than from under-unlearning.
 
\textbf{Dense-Level Utility Recovery.} The bottom row reports dense-architecture baselines from the original papers~\citep{li2024wmdp,shi2024musemachineunlearningsixway} (models in Appendix~\ref{app:eval_protocol}). Our constrained MoE methods recover utility preservation scores of 40.5--42.7 on C4 for KL, NPO and RMU, closing the gap to the unlearning performance on the dense LLMs (44.6), while unconstrained MoE methods may collapse the utility to as low as 4.3.  GRIP also preserves broad capabilities on MoE evaluations, retaining 86--90\% of zero-shot MMLU and 88--89\% of TruthfulQA against 66\% and 69\% for unconstrained baselines (Appendix~\ref{app:General Baseline}). 
 
%, confirming both that the utility penalty previously observed in MoE unlearning is a consequence of routing instability rather than an inherent limitation of sparse architectures and the GRIP architecture can nearly entirely solve this instability. 
 
\textbf{Direct Comparison with SEUF.}
SEUF's soft regularization~\citep{zhang2024seuf} does not overcome the structural bias toward the routing shortcut. At approximately matched forgetting, it underperforms both GRIP variants on every retain-side metric in Table~\ref{tab:main_results}: RS of 0.89--0.92 against GRIP's $\geq 0.94$, and C4 plateauing at 35.1--37.6 against 39.2--42.7 on those same objectives. Soft penalties therefore permit residual router manipulation that diverts retain-set queries to suboptimal experts, which is exactly what a hard constraint forecloses. Appendix~\ref{app:factorial} separates the hard/soft and single/all-expert axes.
 
\textbf{Performance on 14.3B-parameter MoE.}
To ensure a fair comparison and to evaluate whether GRIP generalizes across MoE architectures, we additionally evaluate on the Qwen1.5-MoE-A2.7B-Chat model used in the original SEUF study. We restrict this comparison to WMDP-Cyber with the RMU objective, as these represent SEUF's strongest reported configuration (other benchmarks are omitted because SEUF was not originally evaluated on them). As detailed further in  Appendix~\ref{app:seuf}, GRIP maintains its advantage even on SEUF's native architecture: while achieving comparable forget accuracy (0.25--0.27 vs 0.25), we demonstrate superior retain accuracy (\textbf{0.61}--\textbf{0.63} vs 0.53, a 15--19\% improvement) and routing stability (\textbf{0.94}--\textbf{0.99} vs 0.87, an 8--14\% improvement). These results also confirm that GRIP's geometric constraints do not lose efficacy across different architectures.
 
\subsection{Unlearning-Utility Tradeoffs Across Training}
 
\begin{wrapfigure}{r}{0.45\textwidth}
  \centering
  \vspace{-12pt}
  \includegraphics[width=0.44\textwidth]{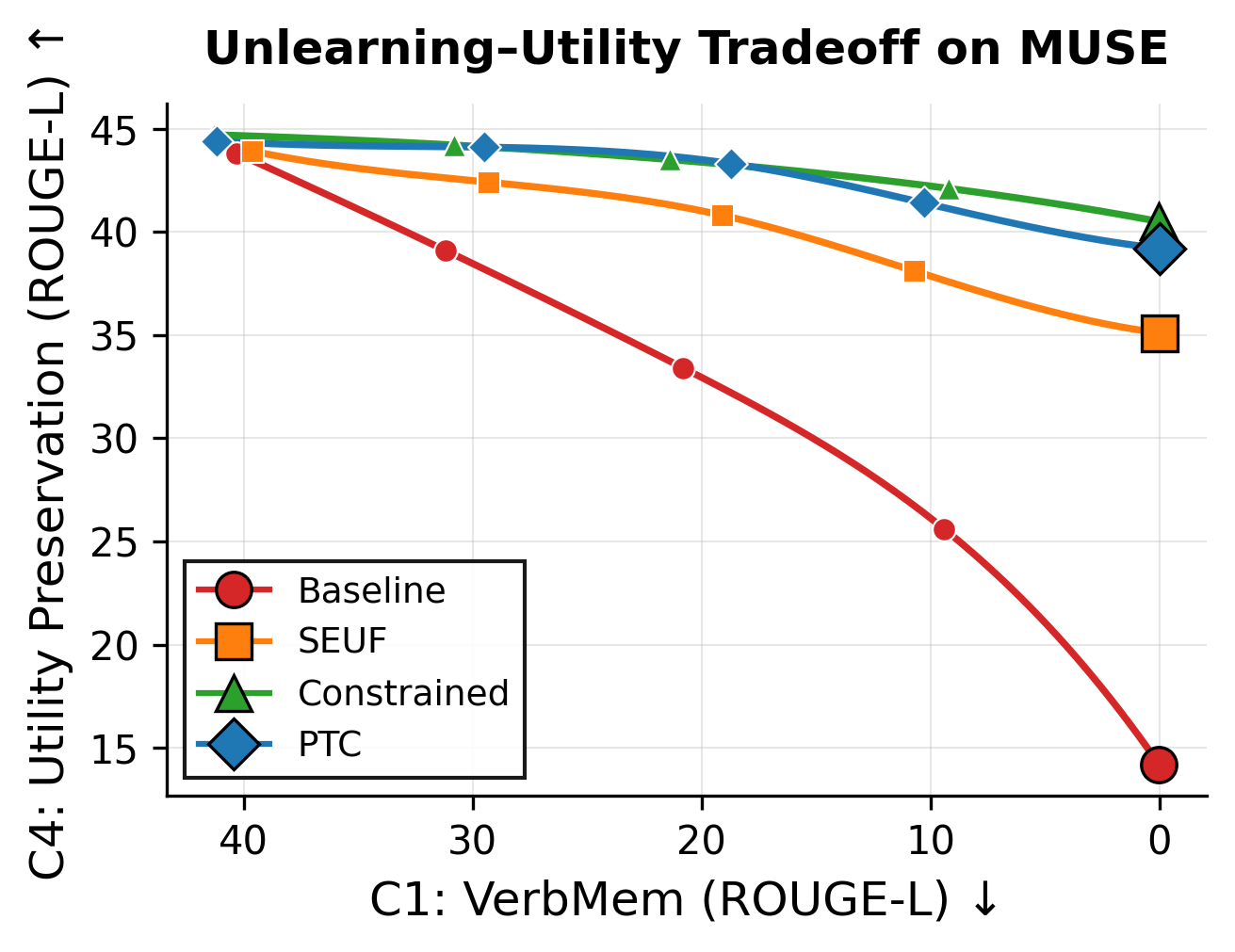}
  \caption{Forget-utility tradeoff on MUSE (RMU). Each curve traces intermediate checkpoints throughout training.}
  \label{fig:tradeoff}
  \vspace{-5pt}
\end{wrapfigure}
 
We track the forget-utility tradeoff across the full unlearning process by evaluating intermediate checkpoints for both unconstrained baselines and GRIP variants. Figure~\ref{fig:tradeoff} plots C4 against C1 metrics at each checkpoint. At every level of forgetting, GRIP-constrained models achieve strictly higher utility than their unconstrained counterparts, confirming that the improvement is not an artifact of early stopping. Unconstrained methods exhibit a steep utility collapse early in training as routing shifts cascade through the network, whereas GRIP maintains a gradual, controlled tradeoff throughout.  We next ablate the individual design choices behind these gains.

\begin{table}[htbp]
    \centering
    \begin{minipage}{0.40\textwidth}
        \centering
        \caption{Analysis of the null-space threshold $\epsilon_{\mathrm{null}}$, distinct from the routing safety margin $\varepsilon$ of Equation~\eqref{eq:batch_feasible}.}
        \label{tab:threshold_sensitivity}
        \small
        \begin{tabular}{l|ccc}
            \toprule
            \textbf{$\epsilon_{\mathrm{null}}$} & \textbf{RS} ($\uparrow$) & \textbf{FA} ($\downarrow$) & \textbf{RA} ($\uparrow$) \\
            \midrule
            $10^{-4}$ & 0.97 & 0.31 & 0.59\\
            $10^{-3}$ & 0.96 & 0.28 & 0.62\\
            $\mathbf{10^{-2}}$ & \textbf{0.96} & \textbf{0.24} & \textbf{0.65} \\
            $10^{-1}$ & 0.89 & 0.25 & 0.59 \\
            \bottomrule
        \end{tabular}
    \end{minipage}
    \hfill 
    \begin{minipage}{0.52\textwidth}
        \centering
        \caption{Ablation analysis of constraint methods. Frozen Router sets $\Delta\Theta_\ell = 0$ and directs all unlearning gradient into the expert MLPs.}
        \label{fig:ablation}
        \small
        \begin{tabular}{l|cccc}
            \toprule
            \textbf{Constraint}& \textbf{FA}$\downarrow$ & \textbf{RA}$\uparrow$  & \textbf{RS}$\uparrow$ & \textbf{Time} $\downarrow$\\
            \midrule
            None (Baseline)& 0.26& 0.35& 0.43 & 1.0$\times$\\
            {Frozen Router}& {0.26}& {0.32}& 0.57 & {1.0$\times$}\\
            Full Null Space& 0.38& 0.68& 0.96 & 1.67$\times$\\
            \textbf{Expert-Specific} & 0.24& 0.65& 0.96 & 1.71$\times$\\
            \textbf{Post-Training}& 0.28& 0.66& 0.99 & 1.21$\times$           \\
            \bottomrule
        \end{tabular}
    \end{minipage}
\end{table}
 
\subsection{Ablation Studies and Design Choices}
\label{sec:ablation}
 
% To validate our design, we conducted an ablation study on different constraint formulations (Table 5) and a sensitivity analysis on the null-space threshold $\epsilon$ (Table 7).
Finally, we conduct ablation analysis on GRIP as well as analysis of null-space thresholds; Appendix~\ref{app:ablations} reports further ablations isolating each design choice. 
 
\textbf{Constraint Formulations.}
As shown in Table~\ref{fig:ablation}, while a strict full 
null-space constraint ($\Delta\Theta X_r = 0$) successfully 
preserves routing ($\text{RS}=0.96$), it overly restricts 
parameter plasticity and degrades unlearning effectiveness 
($\text{FA}=0.38$). Our Expert-Specific formulation resolves 
this trade-off, achieving superior erasure ($\text{FA}=0.24$) 
while maintaining high stability ($\text{RS}=0.96$), though 
at a $1.71\times$ training overhead due to per-step 
projection. Post-Training Correction (PTC) achieves the 
highest stability ($\text{RS}=0.99$) at only $1.21\times$ 
overhead, though its unlearning lags slightly 
($\text{FA}=0.28$) as interim router drift forces expert 
parameters to optimize against a non-stationary routing 
distribution. We recommend the Expert-Specific formulation 
when maximum erasure fidelity is required, and PTC for 
resource-constrained deployments where the 
efficiency--efficacy tradeoff favors throughput.
 
As freezing router \textit{weights} does not freeze routing \textit{decisions}, since routing also depends on the drifting representations reaching each layer (Appendix~\ref{app:frozen}), GRIP's gains do not come merely from immobilizing the router. Table~\ref{fig:ablation} shows that freezing router weights leaves forget efficacy unchanged (FA $=0.26$) but collapses retain accuracy to $0.32$, at baseline level and roughly half of GRIP's $0.65$. 
 
\textbf{Threshold Sensitivity Analysis.}
We sweep the null-space threshold of Appendix~\ref{app:nullspace} across four orders of magnitude ($\epsilon_{\mathrm{null}} \in \{10^{-4}, 10^{-3}, 10^{-2}, 10^{-1}\}$) and use $\epsilon_{\mathrm{null}} = 10^{-2}$ throughout. As shown in Table~\ref{tab:threshold_sensitivity}, tighter values empty the null space in later layers and degrade unlearning ($\text{FA}=0.31$), while looser ones cost routing stability ($0.89$), leaving $10^{-2}$ on the Pareto frontier between stability and plasticity. Appendix~\ref{app:per_layer_eps} resolves this sweep by depth band and finds that the optimum does not shift, though sensitivity to departures from it does; layer-adaptive scheduling remains future work.
 
\subsection{Adversarial Evaluation via Expert Forcing}
\label{sec:adversarial}
The preceding results establish that GRIP preserves model utility while achieving effective forgetting on standard benchmarks. However, these metrics alone cannot distinguish genuine knowledge erasure from retrieval suppression, where targeted information persists in expert parameters but is rendered inaccessible through altered routing paths. Thus, we design an adversarial evaluation that simulates a white-box attacker with full model access who bypasses the post-unlearning router and forces generation through the original pre-unlearning expert selections. Specifically, for forget-set queries that exhibit routing shifts after unlearning, we override the learned router decisions and force the model to use the expert set selected by the pre-unlearning router (full protocol in Appendix~\ref{app:security}), a worst-case diagnostic rather than a deployment threat model (Section~\ref{sec:blackbox}).

\begin{table}[h]
\caption{Expert forcing attack on WMDP. Recovery Rate is 
the fraction of forgotten queries 
recovered when the router is bypassed. GRIP is Method A.}
\centering
\small
\begin{tabular}{lccc}
\toprule
Method & Normal FA$\downarrow$ & Forced FA$\downarrow$ & 
Recovery Rate$\downarrow$ \\
\midrule
Baseline & 0.26 & 0.37 & 11\% \\
SEUF & 0.25 & 0.33 & 8\% \\
\textbf{GRIP} & \textbf{0.24} & \textbf{0.27} & 
\textbf{3\%} \\
\bottomrule
\end{tabular}
\label{tab:expert_forcing}
\end{table}
 
As shown in Table~\ref{tab:expert_forcing}, forced expert access recovers 11\% of supposedly forgotten queries in baseline methods and 8\% under SEUF's soft regularization, whereas GRIP limits recovery to just 3\%, within noise margins. These results confirm that unconstrained and softly-constrained methods exploit routing manipulation as a shortcut that hides knowledge rather than erasing it~\citep{erase_or_hide_2025, unlearning_reversibility_2025}, whereas GRIP's hard geometric constraints force genuine parameter-level erasure that withstands even direct white-box expert access. Further, in order to emulate more realistic attack profiles, we conduct black-box attack analysis on GRIP.
\begin{table}[h]
\caption{Black-box prompt attacks on WMDP-Cyber, following the protocol of \citet{jang2025prompt}. GRIP is Method A.}
\centering
\small
\begin{tabular}{lcccc}
\toprule
Method & Normal FA$\downarrow$ & 5-shot FA$\downarrow$ & Rephrase-as-Poem FA$\downarrow$ & Recovery Rate$\downarrow$ \\
\midrule
Baseline & 0.26 & 0.38 & 0.37 & 11.5\% \\
SEUF & 0.25 & 0.33 & 0.34 & 8.5\% \\
\textbf{GRIP} & \textbf{0.24} & \textbf{0.28} & \textbf{0.29} & \textbf{4.5\%} \\
Dense Baseline & 0.25 & 0.27 & 0.28 & 2.5\% \\
\bottomrule
\end{tabular}
\label{tab:blackbox}
\end{table}
 
\subsection{Black-Box Prompt-Attack Robustness}
\label{sec:blackbox}
 
Following \citet{jang2025prompt}, we run a black-box prompt-attack suite on WMDP-Cyber for the unconstrained baseline, SEUF, and GRIP via \texttt{lm-evaluation-harness}, using two prompt variants and reporting a Recovery Rate analogous to Table~\ref{tab:expert_forcing}. As shown in Table~\ref{tab:blackbox}, GRIP's forget accuracy under attack stays near chance (0.28--0.29) and just above the dense reference, whereas SEUF degrades further (0.33--0.34) and the unconstrained baseline most of all (0.37--0.38), with the Recovery Rate ordering matching. GRIP's recovery is in line with dense-architecture baselines, indicating that it introduces marginal additional black-box vulnerability while additionally closing the white-box expert-forcing loophole the baselines leave open.

\section{Conclusion and Discussion}
\label{sec:conclusion}
 
We identify a structural failure mode in MoE unlearning: standard objectives exploit the router as an optimization shortcut, redirecting forget-set queries away from knowledgeable experts rather than erasing their parameters. GRIP addresses this by projecting router updates into the null space of the retain set's routing matrix, decoupling routing stability from parameter plasticity and forcing erasure to occur within the expert weights themselves. Across two MoE models and two benchmarks, GRIP restores routing stability from 0.21 to at least 0.94, improves retain accuracy by up to 89\%, and reduces white-box adversarial knowledge recovery from 11\% to 3\% while {matching dense-model robustness under black-box prompt attack, at} 1.71$\times$ training overhead, or 1.21$\times$ for the post-training variant. Crucially, GRIP functions as an algorithm-agnostic adapter: it constrains only the router, leaving the underlying unlearning objective unmodified, which means any future dense-architecture unlearning method can be deployed on MoE models by simply wrapping its optimization with GRIP's geometric projection.
 
We acknowledge three limitations. First, caching retain representations incurs $O(LdN_r)$ memory overhead, which may require offloading strategies for extremely deep architectures. Second, GRIP optimizes for faithfulness to the original model's retain-set routing rather than to a retain-only oracle, whose router there is no reason to expect would reproduce the original expert assignments, since the forget data shaped router learning during pretraining; we adopt preservation because the oracle is unattainable at scale and because the preserved routing proves empirically benign for utility, but this assumes the forget set's influence on the global router is small, and the two criteria may diverge for large or heavily retain-entangled forget sets. Preserving the base routing geometry also inherits any biases in the original router; selectively relaxing constraints for biased experts remains an important direction for future work. Third, our claims are empirical rather than formal: we establish no formal guarantee of routing preservation, though the preservation we measure is near-complete (RS $\geq 0.94$).
 
\section*{Acknowledgements}
A. Zhu, R. Wei and P. Li are partially supported by the NSF under awards PHY-2117997, IIS-2239565, IIS-2428777, and CCF-2402816; the JPMorgan Chase Faculty Award; and the OpenAI Researcher Access Program Credit, the Nvidia Academic Award, and the Google Academic Award.
 
\section*{Ethics Statement}
 
This work develops methods for genuine knowledge erasure in MoE language models, motivated by privacy regulations such as GDPR and the need to remove hazardous or copyrighted content from deployed systems. Our experiments use two publicly available benchmarks (WMDP and MUSE) following their original evaluation protocols; we do not generate or distribute hazardous content. The adversarial expert forcing evaluation in Section~\ref{sec:adversarial} demonstrates that existing methods leave sensitive knowledge recoverable, which we disclose to motivate stronger erasure guarantees rather than to provide an exploitation toolkit; the attack requires full white-box model access, limiting its practical applicability.
 
We acknowledge that unlearning methods could in principle be misused to selectively remove safety behaviors from models while preserving harmful capabilities. However, we believe that exposing the superficial nature of current MoE unlearning, where knowledge is hidden by routing manipulation rather than erased, poses the greater risk if left unaddressed, and that publishing principled defenses serves the broader safety community by raising the standard for verifiable knowledge removal.

\bibliography{colm2026_conference}
\bibliographystyle{colm2026_conference}
\appendix
\section{Implementation Details}
\label{sec:appendix_implementation}
 
Our experiments were conducted using the PyTorch framework, leveraging Hugging Face's \texttt{transformers} and \texttt{accelerate} libraries for efficient model handling and distributed training.
 
\textbf{Model Architecture} We use \textbf{Qwen3-30B-A3B}~\citep{qwen3technicalreport}, a Mixture-of-Experts member of the Qwen3 family with $L = 48$ layers, hidden dimension $d = 2048$, and $E = 128$ experts per MoE layer under a top-8 routing strategy. All concrete figures in Appendices~\ref{app:nullspace}--\ref{app:memory} use these values. The model has approximately 3 billion active parameters during inference.
 
\textbf{Hardware and Quantization} All unlearning procedures were executed on a single node equipped with \textbf{2 NVIDIA H200 141GB GPUs} and a single node equipped with \textbf{8 NVIDIA L40S 48GB GPUs}. To maximize throughput and reduce the memory footprint, we employed \textbf{FP8 precision} for both weights and activations during the unlearning steps. Specifically, we utilized the E4M3 floating-point format, which is well-suited for training stability, and integrated NVIDIA's \texttt{TransformerEngine} library to handle the FP8 mixed-precision logic seamlessly.
 
\textbf{Hyperparameters} The unlearning process for our proposed method and all baselines was configured with the hyperparameters listed in Table~\ref{tab:hyperparams}, chosen via a preliminary sweep on a held-out validation set.
 
\begin{table}[h!]
\centering
\caption{Hyperparameters for all unlearning experiments.}
\label{tab:hyperparams}
\begin{tabular}{@{}lc@{}}
\toprule
\textbf{Hyperparameter}     & \textbf{Value}                \\ \midrule
Optimizer                   & AdamW                         \\
Learning Rate               & $3 \times 10^{-5}$              \\
Batch Size (per GPU)        & 2                             \\
Adam $\beta_1$              & 0.9                           \\
Adam $\beta_2$              & 0.999                         \\
Adam $\epsilon$             & $1 \times 10^{-8}$              \\
Weight Decay                & 0.01                          \\
Maximum Unlearning Steps            & 10000                           \\ \bottomrule
\end{tabular}
\end{table}
\subsection{Evaluation Metric Details}
\label{app:eval_protocol}
 
\textbf{WMDP Metrics.}
We follow the evaluation protocol of \citet{li2024wmdp}. 
All models are evaluated on the WMDP-Cyber multiple-choice 
question answering (MCQA) benchmark, which tests knowledge 
of hazardous cybersecurity content.
 
\begin{itemize}
    \item \textbf{Forget Accuracy (FA):} Accuracy on MCQA 
    questions derived from the hazardous cybersecurity 
    corpus. A perfectly unlearned model should perform at 
    random-chance level ($\approx 0.25$ for 4-way MCQA). 
    Lower is better.
    
    \item \textbf{Retain Accuracy (RA):} Accuracy on the 
    MMLU general knowledge MCQA benchmark, measuring 
    preservation of broad model capabilities unrelated to 
    the forget set. Higher is better (approaching the 
    original MoE backbone's score of $\approx 0.70$).
\end{itemize}
 
\noindent\textbf{Controlled Evaluation Protocol.} To 
isolate the effect of routing constraints on utility 
preservation, we control for forgetting level: all methods 
are trained until their forget metrics (FA for WMDP; C1 and 
C2 for MUSE) match the unconstrained baseline as closely as 
the base objective permits, and we compare retain-side 
metrics at that level. This prevents methods from appearing 
superior simply by under-unlearning. We report C3 for 
fidelity to the full MUSE evaluation of 
\citet{shi2024musemachineunlearningsixway} rather than 
because our claims rest on it; being two-sided, it is not 
used as a stopping criterion, and no comparison in this paper 
is drawn from it. The 
match is close for GD, KL and RMU. It is not for NPO: the 
unconstrained baseline and SEUF plateau at FA $0.33$--$0.34$ 
and cannot be driven lower, so the GRIP rows of that block in 
Table~\ref{tab:main_results} are compared against the lowest 
forget accuracy those two methods reach rather than against 
an exactly matched value. The same caveat applies to the 
combined-constraint row of Table~\ref{tab:constraint_side}.
 
\textbf{MUSE Metrics.}
We adhere to the evaluation framework of 
\citet{shi2024musemachineunlearningsixway}, which assesses 
unlearning from both the data owner's and model deployer's 
perspectives. We evaluate on the News corpus. The retrained reference used to normalize C3 is a retain-only finetune of the same backbone under identical hyperparameters, and all C3 values are reported relative to this same-architecture reference rather than to the dense retrained checkpoints released with the benchmark.
 
\begin{itemize}
    \item \textbf{C1: Verbatim Memorization (VerbMem):} 
    The model is prompted with the first $l$ tokens of a 
    sequence from the forget set, and the ROUGE-L F1 score 
    is computed between the model's continuation and the 
    true continuation. This measures the model's tendency to 
    reproduce copyrighted text verbatim. Lower is better.
    
    \item \textbf{C2: Knowledge Memorization (KnowMem):} 
    For each example in the forget set, a question-answer 
    pair $(q, a)$ is constructed. The model generates a 
    response to $q$, and the ROUGE-L score between the 
    generated response and the ground-truth answer $a$ is 
    averaged across all pairs. This measures retention of 
    factual knowledge from the forget set. Lower is better.
    
    \item \textbf{C3: Privacy Leakage (PrivLeak):} 
    Quantifies whether the unlearned model leaks membership 
    information about the forget set. We apply the 
    Min-K\% Prob membership inference attack following the 
    MUSE protocol~\citep{shi2024musemachineunlearningsixway} and compute the AUC-ROC score 
    for distinguishing forget-set members from holdout 
    non-members, normalized against the retrained model's 
    AUC. A value near zero indicates the unlearned model is 
    indistinguishable from a model that was never trained on 
    the forget set; large positive or negative values 
    indicate under- or over-unlearning respectively. Closer 
    to zero is better, so C3 is a two-sided diagnostic rather 
    than a quantity to be minimized; this is why 
    Table~\ref{tab:main_results} marks no best value in that 
    column. Every value we obtain is negative, as is the dense 
    reference, indicating that all methods including the dense 
    baseline sit on the over-unlearning side at the forgetting 
    levels compared here.
    
    \item \textbf{C4: Utility Preservation (KnowMem 
    Retain):} Identical to the C2 metric but evaluated on 
    question-answer pairs derived from the retain set. This 
    measures whether unlearning degrades the model's 
    knowledge of content that should be preserved. Higher is 
    better (approaching the original model's retain-set 
    performance). On the GD objective every method, 
    constrained or not, collapses C4 (4.31--11.4 against 44.6 
    for the dense reference), which is why the C4 comparisons 
    of Section~\ref{sec:results} are scoped to KL, NPO and 
    RMU.
\end{itemize}
 
\noindent\textbf{Routing Stability (RS).} In addition to 
the above benchmark-specific metrics, we report routing 
stability as defined in Equation~\eqref{eq:jaccard} for all 
experiments. RS is computed over the retain set $D_r$ and 
measures the Jaccard similarity between pre- and 
post-unlearning expert selections, averaged across all 
evaluation queries and all MoE layers. An RS of 1.0 
indicates that routing decisions are perfectly preserved; 
values below 0.5 indicate that a substantial fraction of 
expert assignments have changed. Under the top-$8$ routing of 
our backbone an RS of $0.5$ corresponds to roughly three of 
the eight selected experts differing, and half the 
assignments change only at an RS of about $0.33$.
 
\textbf{Dense Baselines.} The bottom row of 
Table~\ref{tab:main_results} reports results from the 
original WMDP and MUSE papers evaluated on dense 
architectures: Zephyr-7B-beta for WMDP 
\citep{li2024wmdp} and LLaMA-2-7B for MUSE 
\citep{shi2024musemachineunlearningsixway}. These serve as 
reference points for the utility ceiling achievable by 
dense-architecture unlearning methods, against which we 
benchmark GRIP's recovery of MoE utility.
 
\subsection{Unlearning Baseline Formulations}
\label{app:baselines}
To ensure fair comparison, we standardized the loss functions for all baseline methods used in Section 5. Let $f_\theta(x)$ denote the model output and $\mathcal{L}_{CE}$ denote the cross-entropy loss.
 
Each objective pairs a forget-side term with a retain-side term of weight $\alpha$. The forget-side term is what distinguishes them, except for KL, which instead substitutes a divergence anchor on the retain side.
 
\textbf{Gradient Difference (GD).} We maximize the loss on the forget set $D_f$ while minimizing loss on the retain set $D_r$ (simulated via gradient accumulation or replay).
\begin{equation}
    \mathcal{L}_{GD} = -\mathcal{L}_{CE}(f_\theta(x_f), y_f) + \alpha \mathcal{L}_{CE}(f_\theta(x_r), y_r)
\end{equation}
 
\textbf{KL Minimization (KL).} We apply the same forget-side ascent, but replace the retain-side cross-entropy with a KL-divergence anchor to a fixed reference model (the initial pre-trained model $\theta_{ref}$), holding retain-set behavior close to the original model rather than only to its labels.
\begin{equation}
    \mathcal{L}_{KL} = -\mathcal{L}_{CE}(f_\theta(x_f), y_f) + \alpha\, \mathrm{KL}\bigl(f_{\theta_{ref}}(x_r) \,\|\, f_\theta(x_r)\bigr)
\end{equation}
 
\textbf{Negative Preference Optimization (NPO).} Following \citet{zhang2024negative}, we treat the forget data as a negative preference sample. The objective drives the forget likelihood below that of the reference model while remaining bounded, avoiding the divergence of unbounded ascent:
\begin{equation}
    \mathcal{L}_{NPO} = -\frac{2}{\beta}\log \sigma \left( -\beta \log \frac{\pi_\theta(y_f|x_f)}{\pi_{ref}(y_f|x_f)} \right) + \alpha \mathcal{L}_{CE}(f_\theta(x_r), y_r)
\end{equation}
where $\beta$ controls the sharpness of the penalty; as $\beta \to 0$ the forget-side term recovers gradient ascent.
 
\textbf{Representation Misdirection (RMU).} Following \citet{li2024wmdp}, we perturb the internal activations $h_\ell(x_f)$ at layer $\ell$ toward a scaled random target vector $v_{rand}$, preventing the extraction of semantic knowledge, while anchoring retain activations to those of the frozen reference model $h^{ref}_\ell$:
\begin{equation}
    \mathcal{L}_{RMU} = \bigl\| h_\ell(x_f) - c\, v_{rand} \bigr\|_2^2 + \alpha \bigl\| h_\ell(x_r) - h^{ref}_\ell(x_r) \bigr\|_2^2
\end{equation}
where $c$ is the activation scaling coefficient of \citet{li2024wmdp}.
 
\subsection{Performance on General Utility Benchmarks}
\label{app:General Baseline}
Beyond the retain-set accuracy reported as RA in Table~\ref{tab:main_results}, which follows the few-shot WMDP harness configuration of \citet{li2024wmdp}, we separately verify that GRIP preserves broad model capabilities by evaluating \textit{zero-shot} performance on MMLU and TruthfulQA. The two MMLU figures are measured under different protocols, so their retention ratios are not directly comparable. GRIP-unlearned models retain 86--90\% of the original model's MMLU score and 88--89\% on TruthfulQA, significantly outperforming unconstrained baselines, which suffer catastrophic utility collapse and retain only around 66\% and 69\% for MMLU and TruthfulQA respectively. Together with the retain-set and routing stability gains above, these results establish that GRIP's constraints preserve model quality at multiple granularities.
 
\section{Null Space Projection Construction}
\label{app:nullspace}
We provide the detailed construction of the null space projector $P^{\mathrm{eq}}_{j,\ell}$ used in Section~\ref{sec:method_a}.
 
\subsection{Eigen-decomposition Method}
Given the retain representations $X^{\mathrm{eq}}_{j,\ell} \in \mathbb{R}^{d \times |\mathcal{I}_{j,\ell}|}$ that selected expert $j$ at layer $\ell$, we compute the covariance matrix:
\begin{equation}
C_{j,\ell} = X^{\mathrm{eq}}_{j,\ell}X^{\mathrm{eq}\top}_{j,\ell} \in \mathbb{R}^{d \times d}
\end{equation}
We perform eigendecomposition of $C_{j,\ell}$:
\begin{equation}
C_{j,\ell} = U_{j,\ell} \Lambda_{j,\ell} U_{j,\ell}^\top
\end{equation}
where $U_{j,\ell} = [u_1, u_2, \ldots, u_d] \in \mathbb{R}^{d \times d}$ contains orthonormal eigenvectors and $\Lambda_{j,\ell} = \text{diag}(\lambda_1, \lambda_2, \ldots, \lambda_d)$ contains eigenvalues ordered as $\lambda_1 \geq \lambda_2 \geq \cdots \geq \lambda_d \geq 0$.
 
The null space of $X^{\mathrm{eq}\top}_{j,\ell}$ corresponds to eigenvectors with zero (or near-zero) eigenvalues. In practice, we use a threshold $\epsilon_{\mathrm{null}} = 10^{-2}$ to identify null space eigenvectors:
\begin{equation}
\hat{U}_{j,\ell} = [u_i : \lambda_i < \epsilon_{\mathrm{null}}] \in \mathbb{R}^{d \times m}
\end{equation}
where $m = |\{i : \lambda_i < \epsilon_{\mathrm{null}}\}|$ is the dimension of the approximate null space. The null space projector is then:
\begin{equation}
P^{\mathrm{eq}}_{j,\ell} = \hat{U}_{j,\ell} \hat{U}_{j,\ell}^\top
\end{equation}
For any router row update $u_{j,\ell} \in \mathbb{R}^d$, the projected update $\tilde{u}_{j,\ell} = P^{\mathrm{eq}}_{j,\ell} u_{j,\ell}$ satisfies $\tilde{u}_{j,\ell}^\top X^{\mathrm{eq}}_{j,\ell} \approx 0$ within numerical precision.
 
\subsection{Parameterization for Unconstrained Optimization}
The null space projector can be used to reparameterize router updates. Instead of optimizing $\Delta\Theta_\ell$ directly with constraints, we can optimize an unconstrained matrix $W_\ell \in \mathbb{R}^{E \times d}$ and set:
\begin{equation}
(\Delta\Theta_\ell)_{j,:} = P^{\mathrm{eq}}_{j,\ell} (W_\ell)_{j,:}
\end{equation}
This ensures each router row update resides in $\text{Null}(X^{\mathrm{eq}\top}_{j,\ell})$ by construction, eliminating the need for explicit constraint enforcement. It enforces only the Stage 1 equality constraints of Section~\ref{sec:method_a}, discarding the Stage 2 freedom to move non-selected scores within their margins, and is therefore more restrictive than the formulation we use, though still less restrictive than the global constraint $\Delta\Theta_\ell X_{r,\ell} = 0$ of Section~\ref{sec:relaxed}.
 
\subsection{Computational Complexity}
The eigendecomposition of $C_{j,\ell} \in \mathbb{R}^{d \times d}$ requires $\mathcal{O}(d^3)$ operations. Because the retain representations drift as unlearning proceeds (Section~\ref{sec:method_b}), the projector is in principle a per-step quantity, and re-forming it from scratch at every step would cost $\mathcal{O}(KLEd^3)$. Our implementation instead performs the full decomposition once per expert and layer before unlearning begins and thereafter edits the cached retain-subspace basis incrementally, so the recurring per-step cost is that of the low-rank update and its application rather than of a fresh decomposition. Storing $\hat{U}_{j,\ell} \in \mathbb{R}^{d \times m}$ requires $\mathcal{O}(dm)$ memory, where $m = d - r_{j,\ell}$ and $r_{j,\ell} = \operatorname{rank}(X^{\mathrm{eq}}_{j,\ell}) \leq |\mathcal{I}_{j,\ell}|$ is the rank of the retain subspace that expert $j$ actually serves. Because top-$k$ routing selects each expert for only a small fraction of retain inputs, $r_{j,\ell} \ll d$ and hence $m$ is close to $d$; Appendix~\ref{app:memory} exploits this by storing the small retain-subspace basis instead.
 
Applying the projection to a router row update $u_{j,\ell} \in \mathbb{R}^d$ requires:
\begin{equation}
\tilde{u}_{j,\ell} = \hat{U}_{j,\ell} (\hat{U}_{j,\ell}^\top u_{j,\ell}) \in \mathcal{O}(dm)
\end{equation}
For our configuration $E=128$, $d=2048$, $m \approx 2000$, this is approximately $4.1 \times 10^6$ operations per expert per projection step.

\section{Randomized Kaczmarz for Halfspace Projection}
\label{app:halfspace}
 
We employ the Randomized Kaczmarz (RK) algorithm to efficiently enforce the inequality constraints described in Section~\ref{sec:method_a}. Unlike standard cyclic projections, RK samples constraints with probability proportional to their geometric ``importance'' (squared Euclidean norms), leading to faster convergence rates.
 
\subsection{Problem Formulation}
 
For expert $j$ at layer $\ell$, we must find an update $\tilde{u}_{j,\ell} \in \mathbb{R}^d$ that satisfies the set of linear inequalities for the non-selected inputs ($i \notin \mathcal{I}_{j,\ell}$):
\begin{equation}
\langle \tilde{u}_{j,\ell}, x^{(i)}_{r,\ell} \rangle \leq \tau_{i,j,\ell} - \varepsilon
\label{eq:inequality_constraint}
\end{equation}
where $\tau_{i,j,\ell}$ is the margin threshold defined in Equation~\eqref{eq:relaxed_constraint}. Let $X_{\mathrm{ineq}}$ be the matrix whose rows are the input vectors $x^{(i)}_{r,\ell}$ for all $i \notin \mathcal{I}_{j,\ell}$, with corresponding thresholds $\tau_{i,j,\ell} - \varepsilon$.
 
While the classic Kaczmarz method sweeps through rows cyclically, the Randomized Kaczmarz algorithm selects a constraint index $i$ at iteration $k$ with probability $p_i$:
\begin{equation}
p_i = \frac{\|x^{(i)}_{r,\ell}\|_2^2}{\|X_{\mathrm{ineq}}\|_F^2}
\end{equation}
where $\|X_{\mathrm{ineq}}\|_F^2 = \sum_{i \notin \mathcal{I}_{j,\ell}} \|x^{(i)}_{r,\ell}\|_2^2$ is the squared Frobenius norm of the constraint matrix.
 
% \section{Randomized Kaczmarz for Halfspace Projection}
% \label{app:halfspace}
 
% We employ the Randomized Kaczmarz (RK) algorithm to efficiently enforce the inequality constraints described in Section~\ref{sec:enforcement}. Unlike standard cyclic projections, RK samples constraints with probability proportional to their geometric "importance" (squared Euclidean norms), leading to faster convergence rates.
 
% \subsection{Problem Formulation}
 
% For expert $j$ at layer $\ell$, we must find a gradient update $\tilde{\nabla}$ that satisfies the set of linear inequalities for the non-selected inputs ($i \notin \mathcal{I}_{j,\ell}$):
% \begin{equation}
% \langle \tilde{\nabla}, x^{(i)}_{r,\ell} \rangle \leq \tau_{i,j} - \varepsilon
% \end{equation}
% where $\tau_{i,j}$ is the margin threshold. Let $A$ be the matrix where rows correspond to the input vectors $x^{(i)}_{r,\ell}$ for all $i \notin \mathcal{I}_{j,\ell}$, and let $b$ be the vector of thresholds $\tau_{i,j} - \varepsilon$.
 
% While the classic Kaczmarz method sweeps through rows cyclically, the Randomized Kaczmarz algorithm selects a constraint index $i$ at iteration $k$ with probability $p_i$:
% \begin{equation}
% p_i = \frac{\|x^{(i)}_{r,\ell}\|_2^2}{\|A\|_F^2}
% \end{equation}
% where $\|A\|_F^2 = \sum_{m} \|x^{(m)}_{r,\ell}\|_2^2$ is the Frobenius norm of the constraint matrix.
 
\subsection{Algorithm}
 
Algorithm~\ref{alg:training_rk} details the procedure. We precompute the row norms and sampling probabilities. At each step, we sample a constraint; if the constraint is violated, we project the current router row update orthogonally onto the defining hyperplane.
 
\begin{algorithm}[h!]
\caption{GRIP Training-Time Enforcement via Randomized Kaczmarz}
\label{alg:training_rk}
\begin{algorithmic}[1]
\REQUIRE Unconstrained router row updates $\{u_{j,\ell}\}$, retain representations $X_{r,\ell}$, expert selections $\{\mathcal{I}_{j,\ell}\}$, thresholds $\{\tau_{i,j,\ell}\}$
\ENSURE Constrained updates $\{\bar{u}_{j,\ell}\}$
 
\FOR{each layer $\ell = 1, \dots, L$}
    \FOR{each expert $j = 1, \dots, E$}
        \STATE \COMMENT{--- Phase 1: Equality Constraints (Null Space) ---}
        \STATE $X^{\mathrm{eq}}_{j,\ell} \gets [x^{(i)}_{r,\ell}]_{i \in \mathcal{I}_{j,\ell}}$
        \STATE $P^{\mathrm{eq}}_{j,\ell} \gets I - X^{\mathrm{eq}}_{j,\ell} ({X^{\mathrm{eq}}_{j,\ell}}^\top X^{\mathrm{eq}}_{j,\ell})^{\dagger} {X^{\mathrm{eq}}_{j,\ell}}^\top$ \COMMENT{cached, updated incrementally (App.~\ref{app:nullspace})}
        \STATE $\tilde{u}^{(0)}_{j,\ell} \gets P^{\mathrm{eq}}_{j,\ell} u_{j,\ell}$
        
        \STATE \COMMENT{--- Phase 2: Inequality Constraints (Randomized Kaczmarz) ---}
        \STATE Let $X_{\mathrm{ineq}}$ be the matrix of rows $x^{(i)}_{r,\ell}$ for $i \notin \mathcal{I}_{j,\ell}$
        \STATE Compute row norms $w_i = \|x^{(i)}_{r,\ell}\|_2^2$
        \STATE Compute probabilities $p_i = w_i / \sum_{i' \notin \mathcal{I}_{j,\ell}} w_{i'}$
        \STATE $k \gets 0$
        
        \WHILE{$k < k_{\max}$}
            \STATE Sample index $i$ from distribution $p$
            \STATE $v_{i,j,\ell} \gets (\tilde{u}^{(k)}_{j,\ell})^\top x^{(i)}_{r,\ell} - (\tau_{i,j,\ell} - \varepsilon)$ \COMMENT{Compute Residual}
            \IF{$v_{i,j,\ell} > 0$}
                \STATE $\tilde{u}^{(k+1)}_{j,\ell} \gets \tilde{u}^{(k)}_{j,\ell} - \frac{v_{i,j,\ell}}{w_i} \cdot x^{(i)}_{r,\ell}$
            \ELSE
                \STATE $\tilde{u}^{(k+1)}_{j,\ell} \gets \tilde{u}^{(k)}_{j,\ell}$
            \ENDIF
            \STATE $k \gets k + 1$
            \STATE \COMMENT{Optional: Early exit if converged}
        \ENDWHILE
        \STATE $\bar{u}_{j,\ell} \gets \tilde{u}^{(k)}_{j,\ell}$
    \ENDFOR
\ENDFOR
\RETURN $\{\bar{u}_{j,\ell}\}$
\end{algorithmic}
\end{algorithm}
 
\subsection{Convergence Analysis}
 
The classical randomized Kaczmarz analysis of \citet{strohmer2009randomized} concerns \textit{consistent linear equality systems}, whereas $\mathcal{K}_{j,\ell}$ is a mixed system of equalities and inequalities; that result therefore does not by itself cover the projection we perform. \citet{leventhal2010randomized} extend randomized iterated projection precisely to systems of linear equalities and inequalities, bounding the linear rate of convergence in expectation by the Hoffman constant of the system, and the Sampling Kaczmarz--Motzkin family \citep{deloera2017sampling} establishes comparable rates for linear feasibility. It is those results we rely on, and because their analysis admits mixed systems, a single statement covers both stages of our procedure rather than Stage 2 alone.
 
{We state the guarantee as a \textit{per-step} claim. Within a single optimization step the retain representations $X_{r,\ell}$, active-expert sets $\mathcal{I}_{j,\ell}$, and margins $\tau_{i,j,\ell}$ are fixed, so $\mathcal{K}_{j,\ell}$ in Equation~\eqref{eq:batch_feasible} is a fixed intersection of half-spaces and the system solved is a fixed, consistent inequality system, exactly the setting the cited results assume.}
 
\begin{theorem}[Per-step convergence of the Stage 2 projection]
\label{thm:rk}
Fix an optimization step, holding $X_{r,\ell}$, $\mathcal{I}_{j,\ell}$, and $\{\tau_{i,j,\ell}\}$ fixed, so that $\mathcal{K}_{j,\ell}$ is a fixed nonempty polyhedron. Then the iterates of Algorithm~\ref{alg:training_rk} satisfy
\begin{equation}
\mathbb{E} \bigl[ \operatorname{dist}\bigl(\tilde{u}^{(k)}_{j,\ell},\, \mathcal{K}_{j,\ell}\bigr)^2 \bigr]
\leq \Bigl( 1 - \tfrac{1}{H^2 \|X_{\mathrm{ineq}}\|_F^2} \Bigr)^{k}
\operatorname{dist}\bigl(\tilde{u}^{(0)}_{j,\ell},\, \mathcal{K}_{j,\ell}\bigr)^2
\end{equation}
where $\operatorname{dist}(\cdot, \mathcal{K}_{j,\ell})$ is the Euclidean distance to the feasible set, $X_{\mathrm{ineq}}$ collects the retain representations defining the constraints, and $H$ is the Hoffman constant of that system, written $H$ rather than $L$ to avoid collision with the layer count.
\end{theorem}

\begin{remark}[Feasible point, not nearest point]
The bound is on the distance to $\mathcal{K}_{j,\ell}$, not to any particular element of it. Unlike the equality case, where randomized Kaczmarz converges to the Euclidean projection of the starting iterate onto the solution set, projection onto an intersection of half-spaces converges to \textit{some} feasible update, which need not be the projection of $\tilde{u}^{(0)}_{j,\ell}$. The procedure therefore returns a router update that satisfies the routing constraints, not the minimum-norm modification of the base algorithm's update; we do not rely on the latter anywhere in the method.
\end{remark}
 
\begin{remark}[Scope of Theorem~\ref{thm:rk}]
The statement is within-step only. Across optimization steps the feasible region itself moves, since upstream updates shift $X_{r,\ell}$ and hence both the active sets and the margins, so the procedure solves a sequence of related but distinct feasibility problems. We do not claim the per-step rate transfers to this non-stationary setting without further conditions, and we make no formal claim about the outer optimization trajectory; this is part of what motivates Method B, which defers enforcement until the representations have stopped moving.
\end{remark}
 
Within a step, the linear rate $\mathcal{O}(c^k)$, as against the $\mathcal{O}(1/\sqrt{k})$ behavior of cyclic projections, is what allows a small fixed iteration budget ($k_{\max} \approx 100$) to drive the residual violation low even when the number of retain inputs $N_r$ is large.
 
\subsection{Computational Complexity}
 
The computational cost is dominated by the probability precomputation and the iterative updates:
\begin{itemize}
    \item \textbf{Precomputation:} Computing squared norms for all non-selected inputs takes $O(d \cdot N_{\text{ineq}})$. This is done once per batch.
    \item \textbf{Per Iteration:} Sampling takes $O(\log N_{\text{ineq}})$ or $O(1)$ with aliasing methods. The update step involves a single dot product and vector addition: $O(d)$.
    \item \textbf{Total Complexity:} $O(d N_{\text{ineq}} + k_{\max} d)$.
\end{itemize}
 
Since $k_{\max}$ is a small constant independent of the system size (due to the linear convergence rate), the marginal cost of enforcement scales linearly with the dimension $d$, which is computationally efficient at the router hidden size of our backbone ($d = 2048$).
\subsection{Time Complexity of PTC versus Training-Time Enforcement}
 
\textbf{Time Complexity Efficiency.} 
The computational advantage of Post-Training Correction (PTC) over Training-Time Enforcement (TTE) stems from the frequency of projection operations:
\begin{itemize}
    \item \textbf{TTE Cost:} Requires projecting updates at every one of the $K$ unlearning steps, for every layer and expert. Re-forming the projectors would scale as $O(K \cdot L \cdot E \cdot d^3)$ with $K \approx 10{,}000$; the incremental scheme of Appendix~\ref{app:nullspace} avoids the repeated decomposition, but a per-step projection cost remains.
    \item \textbf{PTC Cost:} Requires a single analytical correction per layer after training. Total cost scales as $O(1 \cdot L \cdot d^3)$.
\end{itemize}
Since $K \gg 1$, PTC reduces the projection term by approximately four orders of magnitude. That term is not the dominant cost of an unlearning step in absolute terms, since backpropagation through the model and the optimizer update are larger; it is dominant only within the overhead that constraint enforcement adds. Removing it therefore does not return the total to $1.0\times$, but lowers it from $1.71\times$ to the $1.21\times$ reported in Section~\ref{sec:method_b}, the residual being representation extraction and the FP32 solve.
 
\section{Efficient Implementation and Storage}
\label{app:memory}
 
Enforcing geometric constraints on large-scale Mixture-of-Experts (MoE) models presents a unique systems challenge. While the parameter updates are sparse, the constraint matrices are dense, layer-specific, and expert-specific. Storing a dense $d \times d$ projector for every expert across all layers would require prohibitive memory. For our 30B backbone ($d=2048$, $E=128$, $L=48$), naive storage would require $d^2 E L \approx 2.6 \times 10^{10}$ entries, roughly 52~GB in FP16 and over 100~GB in FP32. To resolve this, we implement a \textit{Tiered Hierarchical Caching Strategy}, reducing the active memory footprint to $<1\%$ of the model size without stalling the training pipeline.
 
\subsection{Implicit Low-Rank Storage}
Instead of instantiating and storing the full dense projector $P^{\mathrm{eq}}_{j,\ell} \in \mathbb{R}^{d \times d}$, we exploit the spectral properties of the router representations. Because top-$k$ routing selects any given expert for only a small fraction of retain inputs, the retain subspace that expert $j$ serves is effectively low-rank: $r_{j,\ell} \ll d$, where $r_{j,\ell} = \operatorname{rank}(X^{\mathrm{eq}}_{j,\ell})$. We therefore store the orthonormal basis of that subspace, $U^{\mathrm{ret}}_{j,\ell} \in \mathbb{R}^{d \times r_{j,\ell}}$, rather than the much larger null-space basis, and apply the projector implicitly through its complement form:
\begin{equation}
\label{eq:implicit_proj}
P^{\mathrm{eq}}_{j,\ell}\, u_{j,\ell} = u_{j,\ell} - U^{\mathrm{ret}}_{j,\ell} \bigl( U^{\mathrm{ret}\top}_{j,\ell} u_{j,\ell} \bigr)
\end{equation}
Three expressions for the same operator appear in this paper and are worth separating. The pseudo-inverse form of Section~\ref{sec:method_a} is the exact projector, against which the method is defined. The thresholded eigendecomposition of Appendix~\ref{app:nullspace} is how that projector is computed, and coincides with it up to the truncation at $\epsilon_{\mathrm{null}}$, which is the sole discrepancy between the two and the subject of Table~\ref{tab:threshold_sensitivity}. Equation~\eqref{eq:implicit_proj} is how the computed projector is stored and applied, and is algebraically identical to the eigendecomposition form since $U^{\mathrm{ret}}_{j,\ell}$ and $\hat{U}_{j,\ell}$ span complementary subspaces.
 
This reduces the storage complexity per expert from $\mathcal{O}(d^2)$ to $\mathcal{O}(d\,r_{j,\ell})$. For a typical rank $r_{j,\ell} \approx 50$ and hidden dimension $d=2048$, this yields a compression ratio of $d / r_{j,\ell} \approx 40\times$, allowing the structural constraints for thousands of experts to reside efficiently in system RAM or on disk.
 
\subsection{Hierarchical Caching System}
To manage the flow of these constraints during the unlearning optimization, we developed a custom PyTorch autograd hook (\texttt{MemoryEfficientConstraintHook}) that implements a three-tier Least Recently Used (LRU) cache replacement policy:
\begin{itemize}
    \item \textbf{Tier 1: Hot Cache (VRAM).} Stores the retain-subspace bases $U^{\mathrm{ret}}_{j,\ell}$ for the currently active experts in GPU memory, applied in the implicit form of Equation~\eqref{eq:implicit_proj} rather than materialized as dense $d \times d$ projectors. Capacity is strictly limited (e.g., $C_{\text{hot}}=8$ experts) to prevent out-of-memory (OOM) errors during the backward pass.
    \item \textbf{Tier 2: Warm Cache (Host RAM).} Stores compressed basis vectors $U^{\mathrm{ret}}_{j,\ell}$ for frequently accessed experts in CPU memory. This tier serves as a low-latency buffer, avoiding the I/O bottleneck of disk reads.
    \item \textbf{Tier 3: Cold Storage (NVMe/Disk).} The complete set of expert constraints is sharded into individual files on disk. These are loaded lazily only when a ``cache miss'' occurs in the upper tiers.
\end{itemize}
 
\subsection{Stochastic Fallback Mechanism}
To ensure numerical stability without incurring unnecessary computation, our implementation employs an adaptive verification step. During the backward pass, the hook first applies the Stage 1 equality projection. It then checks the residual violation against the inequality constraints. Only if the violation exceeds a numerical tolerance ($\eta > 10^{-4}$, distinct from the selection margins $\tau_{i,j,\ell}$) does the system trigger the Randomized Kaczmarz solver. This lazy enforcement ensures that the iterative projection is only invoked for the small subset of gradients that actually threaten the routing stability bounds.
 
\subsection{Post-Training Correction: Complexity and Memory Analysis}
\label{app:complexity}
 
We provide a detailed complexity analysis of the Post-Training 
Correction (PTC) procedure described in 
Section~\ref{sec:method_b}, covering both computational cost 
and memory management.
 
\paragraph{Time Complexity.}
The dominant cost per layer is the $O(d^3)$ linear solve of 
Equation~\eqref{eq:correction}. Across all $L$ MoE layers, the total PTC cost is 
$O(L \cdot d^3)$. In contrast, Training-Time Enforcement 
(Section~\ref{sec:method_a}) requires a projection at every 
one of the $K$ unlearning steps and for every expert, 
yielding $O(K \cdot L \cdot E \cdot d^3)$ if projectors are 
re-formed rather than updated incrementally. Since 
$K \approx 10{,}000$ in our experiments, PTC reduces the 
projection term by approximately four orders of magnitude. 
That term is not the dominant cost of training as a whole, 
which is set by backpropagation and the optimizer; it is the 
dominant part of the overhead GRIP adds, which is why the 
reduction appears as $1.71\times \to 1.21\times$ rather than 
as a return to $1.0\times$.
 
\paragraph{Numerical Precision.}
All intermediate computations in the correction are performed 
in FP32 regardless of the model's native precision (FP8 E4M3 
in our experiments). This is necessary because the Gram 
matrix $\tilde{X}_{r,\ell}\tilde{X}_{r,\ell}^{\top}$ and linear solve are numerically sensitive: the 
3-bit mantissa of E4M3 introduces unacceptable accumulation 
error in the $O(d^2 N_r)$ outer product. The ridge term $\lambda = 10^{-6}$ of 
Equation~\eqref{eq:correction} is applied in this precision, 
sized to regularize the solve without perceptibly shifting 
the restored scores; the acceptance check below verifies this 
per layer. The 
final correction $\Delta_\ell$ is cast back to the 
model's native dtype before application. We verify each 
layer's correction by computing the cosine similarity between 
original routing scores $\Theta_\ell X_{r,\ell}$ and 
corrected scores $(\tilde{\Theta}_\ell + \Delta_\ell) \tilde{X}_{r,\ell}$, requiring similarity $> 0.95$ for acceptance. 
As noted in Section~\ref{sec:method_b}, this check doubles as 
the detector for PTC's failure mode: a layer that fails it 
should fall back to Method A.
 
\phantomsection
\paragraph{Memory Footprint.}
~\label{app:ptc_memory}
The figures below assume cached tensors are held in FP16 rather than in
the model's native FP8, and the correction arithmetic in FP32. This is a
hardware limitation rather than a design choice: E4M3 is exposed only
through the TransformerEngine compute path on our hardware, and the
host-RAM and disk tiers used for offloading provide no native FP8 storage
format, so cached representations and router copies are upcast on
eviction. The cache is therefore larger than the model's own precision
would suggest, and the numbers below are the values that actually apply
in our setup.
 
PTC's memory usage is dominated by three components:
 
\begin{enumerate}
    \item \textbf{Retain representations} 
    ($2 \times L \times d \times N_r$): Storing both 
    $X_{r,\ell}$ and $\tilde{X}_{r,\ell}$ across all layers. For 
    our 30B model ($L = 48$, $d = 2048$, and retain-set size 
    $N_r = 2048$) in FP16, this totals approximately 
    $2 \times 48 \times 2048 \times 2048 \times 2\text{B} 
    \approx 0.8$\,GB. However, 
    because PTC processes layers sequentially, only one 
    layer's representations need reside in VRAM at a time; 
    the remainder are offloaded to CPU RAM.
    
    \item \textbf{Working memory} ($O(d^2)$): The Gram 
    matrix $\tilde{X}_{r,\ell}\tilde{X}_{r,\ell}^{\top} \in \mathbb{R}^{d \times d}$ of 
    Equation~\eqref{eq:correction} and the linear 
    solve workspace require $2048^2 \times 4\text{B} \approx 
    17$\,MB in FP32 per layer, allocated and freed within 
    each layer's computation.
    
    \item \textbf{Original router weights} ($L \times E 
    \times d$): A copy of each layer's pre-unlearning gate 
    weights $\Theta_\ell \in \mathbb{R}^{E \times d}$ is 
    required to compute the correction target. For $E = 128$ 
    experts, this totals $48 \times 128 \times 2048 \times 
    2\text{B} \approx 25$\,MB in FP16.
\end{enumerate}
 
\paragraph{Caching and Amortization.}
Because the original model's representations $X_{r,\ell}$ 
and gate weights $\Theta_\ell$ are independent of the 
unlearning objective, we cache them to disk after a single 
extraction pass. Subsequent PTC runs (e.g., with different 
unlearning algorithms or hyperparameters) skip the original 
model entirely, requiring only a forward pass through the 
unlearned model to obtain $\tilde{X}_{r,\ell}$. Cache integrity is 
validated via metadata checks on the model identifier, 
retain-set size, sample fingerprint, and layer indices; 
mismatches trigger automatic recomputation. Individual 
layer corrections $\Delta_\ell$ are also cached to 
disk, enabling incremental recomputation when only a 
subset of layers requires updating. In practice, this 
caching reduces the wall-clock time of repeated PTC 
applications from $\sim$45 minutes (with original model 
loading and extraction) to $\sim$8 minutes (unlearned model 
extraction and correction only) on our hardware 
configuration (2$\times$ H200 141GB).
\section{Security Evaluation Protocols}
\label{app:security}
 
To rigorously quantify the robustness of unlearning, we define the formal evaluation algorithm \textbf{Adversarial Expert Forcing}, which tests for latent knowledge retention in expert parameters.
 
\subsection{Adversarial Expert Forcing}
This protocol simulates a white-box adversary who bypasses the router's suppression mechanism to directly access erased knowledge; Algorithm~\ref{alg:expert_forcing} states the procedure. Two quantities must be distinguished. The \textit{Adversarial Recovery Rate} (ARR) returned by the algorithm is the raw accuracy of the unlearned model when forced onto the pre-unlearning routing path. The \textit{Recovery Rate} of Table~\ref{tab:expert_forcing} is its increase over the model's accuracy under its own routing, $\mathrm{ARR} - \mathrm{FA}_{\text{normal}}$, which isolates what forcing recovers rather than what the model answers correctly regardless. The Recovery Rate of Table~\ref{tab:blackbox} is the analogous increase, averaged over the two prompt variants of the suite. The query set $\mathcal{D}_{\text{haz}}$ is restricted to forget-set queries whose expert selection actually shifts during unlearning, since forcing is a no-op on the remainder. Both tables report the training-time variant (Method A).
 
\begin{algorithm}[h!]
   \caption{Adversarial Expert Forcing Evaluation}
   \label{alg:expert_forcing}
\begin{algorithmic}[1]
   \STATE {\bfseries Input:} Pre-unlearning Model $\Theta_{pre}$, Unlearned Model $\Theta_{post}$, Hazardous Query Set $\mathcal{D}_{haz}$
   \STATE {\bfseries Input:} Normal-routing forget accuracy $\mathrm{FA}_{\text{normal}}$ of $\Theta_{post}$
   \STATE {\bfseries Output:} ARR, and Recovery Rate $\mathrm{ARR} - \mathrm{FA}_{\text{normal}}$
   \STATE $N_{correct} \leftarrow 0$
   
   \FOR{each input pair $(x, y)$ in $\mathcal{D}_{haz}$}
       \STATE \COMMENT{Step 1: Extract original routing path}
       \STATE $\mathcal{S}_{orig} \leftarrow \text{GetActiveExperts}(\Theta_{pre}, x)$
       
       \STATE \COMMENT{Step 2: Forward pass with forced routing}
       \STATE \COMMENT{Override $\Theta_{post}$ router to strictly use $\mathcal{S}_{orig}$}
       \STATE $\hat{y} \leftarrow \text{Forward}(\Theta_{post}, x, \text{force\_experts}=\mathcal{S}_{orig})$
       
       \STATE \COMMENT{Step 3: Measure recovery}
       \IF{$\text{argmax}(\hat{y}) = y$}
           \STATE $N_{correct} \leftarrow N_{correct} + 1$
       \ENDIF
   \ENDFOR
   
   \STATE $\mathrm{ARR} \gets N_{correct} / |\mathcal{D}_{haz}|$
   \STATE \textbf{return} $\mathrm{ARR}$, and the reported Recovery Rate $\mathrm{ARR} - \mathrm{FA}_{\text{normal}}$
\end{algorithmic}
\end{algorithm}

\section{Direct Comparison with SEUF}
\label{app:seuf}
 
To evaluate the efficacy of our framework against the current state-of-the-art in MoE unlearning, we perform a direct head-to-head comparison with \textbf{SEUF}~\citep{zhang2024seuf}. 
 
\textbf{Experimental Setup.} 
To ensure a strictly fair comparison, we replicate the specific experimental conditions used in the original SEUF study. We utilize the \textbf{Qwen1.5-MoE-A2.7B-Chat} architecture instead of the 30B model used in our main experiments. We evaluate performance on the \textbf{WMDP-Cyber} dataset, consistent with the hazardous knowledge removal tasks targeted by both frameworks. We compare both methods when applied to the Representation Misdirection (RMU) unlearning algorithm.
\begin{table}[h!]
    \caption{Direct comparison with SEUF on Qwen1.5-MoE-A2.7B-Chat. $\downarrow$/$\uparrow$ indicates lower/higher is better. Our results are \textbf{bolded}.}
    \label{tab:seuf_comparison}
    \small
    \centering
    \begin{tabular}{l|ccc}
        \toprule
        \textbf{Method} & FA$\downarrow$ & RA$\uparrow$ & RS $\uparrow$\\
        \midrule
        RMU Baseline & 0.28 & 0.38 & 0.35 \\
        SEUF + RMU & 0.25 & 0.53 & 0.87 \\
        \midrule
        \textbf{Expert-Specific (GRIP) + RMU} & \textbf{0.25} & \textbf{0.61} & \textbf{0.94} \\
        \textbf{PTC (GRIP) + RMU} & 0.27 & \textbf{0.63} & \textbf{0.99}\\
        \bottomrule
    \end{tabular}
\end{table}
 
\subsection{Results and Analysis}
The quantitative results are presented in Table~\ref{tab:seuf_comparison}. Our analysis highlights three key findings:
 
\textbf{Unlearning Efficacy (FA):} Both frameworks successfully reduce the model's accuracy on the forget set to near-random levels. GRIP (Relaxed Constraint) achieves an FA of 0.25, matching SEUF exactly, while PTC achieves 0.27. This confirms that restricting the router's optimization landscape via null-space projections does not hinder the model's ability to erase hazardous knowledge.
    
\textbf{Superior Utility Preservation (RA):} GRIP demonstrates a significant advantage in retaining general model capabilities. While SEUF recovers a Retain Accuracy (RA) of 0.53, GRIP improves this to \textbf{0.61} (+15\%) with relaxed constraints and \textbf{0.63} (+19\%) with PTC. SEUF relies on soft regularization penalties ($L_2$ norms on router weights) to maintain stability. Our results suggest these soft penalties are insufficient to prevent the "optimization shortcut" described in Section 1, leading to degraded utility. In contrast, GRIP's hard geometric constraints strictly enforce the preservation of retain-set routing, protecting general knowledge more effectively.
    
\textbf{Routing Stability (RS):} We observe that SEUF improves routing stability over the unconstrained baseline (0.87 vs 0.35) but still permits noticeable drift. GRIP eliminates this drift almost entirely, achieving an RS of \textbf{0.94} with training-time constraints and \textbf{0.99} with Post-Training Correction. This validates our hypothesis that decoupling routing stability from parameter plasticity is essential for robust MoE unlearning.

\section{Additional Ablations}
\label{app:ablations}
 
This appendix isolates the individual contributions of GRIP's design
choices. Appendix~\ref{app:frozen} reports the frozen-router baseline;
Appendix~\ref{app:factorial} separates the hard-versus-soft
constraint axis from the single-versus-all-expert update axis;
Appendix~\ref{app:constraint_side} compares constraining routing on the
retain set against forget-side and combined alternatives. All runs use WMDP-Cyber with the RMU objective on Qwen3-30B-A3B, at the
approximately matched forget accuracy of Section~\ref{sec:results}; where a
configuration cannot be driven to that level, we report the level it reaches
and say so in the text.
 
\subsection{Frozen Router Baseline}
\label{app:frozen}
 
The most direct baseline against the routing-shortcut hypothesis is to
freeze the router entirely, setting $\Delta\Theta_\ell = 0$ and directing
all unlearning gradient into the expert MLPs. If immobilizing the router
were sufficient, the expert-specific projections and post-hoc correction
of Section~\ref{sec:method} would not be justified. As reported in
Table~\ref{fig:ablation}, forget efficacy is unchanged (FA $= 0.26$,
against $0.26$ for the unconstrained baseline), but utility collapses:
retain accuracy falls to $0.32$, at the level of the unconstrained
baseline ($0.35$) and roughly half of GRIP's $0.65$.
 
The explanation is that freezing router \textit{weights} does not freeze
routing \textit{decisions}. Expert selection depends on both the router
parameters $\Theta_\ell$ and the representations $X_{r,\ell}$ arriving at
layer $\ell$, and as Section~\ref{sec:method_b} describes, those
representations drift ($X_{r,\ell} \to \tilde{X}_{r,\ell}$) as upstream
experts are updated. A frozen-router model therefore still loses routing
stability on the retain set, through upstream drift rather than through
router updates, and unlike the post-training correction of
Equation~\eqref{eq:correction} it has no mechanism for realigning to the
drifted representations. 

\subsection{Factorial Ablation: Constraint Hardness and Update Breadth}
\label{app:factorial}
 
GRIP differs from SEUF~\citep{zhang2024seuf} along two axes at once: it
replaces a soft $L_2$ penalty on router weights with a hard null-space
projection, and it leaves all activated experts free to update rather
than restricting updates to the single highest-attributed expert. The
comparison in Table~\ref{tab:main_results} varies both together, so it
cannot attribute the gain to either. We therefore run the full $2 \times 2$
design, in which the soft/single cell reproduces SEUF and the hard/all
cell is GRIP.
 
\begin{table}[h]
\centering
\caption{Factorial ablation on WMDP-Cyber (RMU). Constraint: soft $L_2$
penalty on router weights versus hard null-space projection. Updates:
single highest-attributed expert versus all activated experts. The
soft/single configuration corresponds to SEUF and hard/all to GRIP; those
two rows reproduce Tables~\ref{tab:main_results} and~\ref{tab:expert_forcing}.
Rec.\ is the forced-recovery rate under Expert Forcing
(Section~\ref{sec:adversarial}), which is the metric that distinguishes
update breadth, since FA is controlled by construction.}
\label{tab:factorial}
\small
\begin{tabular}{llcccc}
\toprule
\textbf{Constraint} & \textbf{Updates} & \textbf{FA}$\downarrow$ & \textbf{RA}$\uparrow$ & \textbf{RS}$\uparrow$ & \textbf{Rec.}$\downarrow$ \\
\midrule
Soft ($L_2$)            & Single expert & 0.25 & 0.58 & 0.89 & 8\% \\
Soft ($L_2$)            & All experts   & 0.26 & 0.47 & 0.88 & 7\% \\
Hard (projection)       & Single expert & 0.27 & 0.60 & 0.96 & 7\% \\
\textbf{Hard (projection)} & \textbf{All experts} & \textbf{0.24} & \textbf{0.65} & \textbf{0.96} & \textbf{3\%} \\
\bottomrule
\end{tabular}
\end{table}

In Table~\ref{tab:factorial}, the two diagonal cells are already
reported: soft/single (SEUF) reaches
RS $0.89$, RA $0.58$ and $8\%$ forced recovery, while hard/all (GRIP)
reaches RS $0.96$, RA $0.65$ and $3\%$. The two off-diagonal cells
isolate which axis is responsible. Holding update breadth fixed, moving
from a soft penalty to a hard projection raises routing stability from
$0.88$--$0.89$ to $0.96$ in both rows. Holding constraint hardness fixed,
broadening updates from a single expert to all activated experts cuts
forced recovery from $7\%$ to $3\%$ under the hard constraint while
leaving routing stability unchanged at $0.96$. The two axes therefore
contribute distinctly: constraint hardness is responsible for routing
stability and the retain-side utility that follows from it, whereas
update breadth is responsible for erasure depth, which forget accuracy
alone cannot reveal because it is matched by construction across all four
cells.
 
The two effects are not independent. Broadening updates under a soft
penalty degrades retain accuracy ($0.58$ to $0.47$) and barely improves
recovery ($8\%$ to $7\%$), because updating all activated experts
amplifies exactly the routing drift a soft penalty fails to prevent.
Update breadth pays off only once routing is held fixed by a hard
constraint, which is the configuration GRIP occupies.

\subsection{Choice of Constraint Surface}
\label{app:constraint_side}
 
Section~\ref{sec:method} constrains routing on the retain set. Two
alternatives are available: constraining forget-set routing, and
constraining both. We compare all three at approximately matched forget accuracy. Note
that this axis is distinct from the granularity axis of
Table~\ref{fig:ablation}, which varies how tightly the retain-side
constraint is enforced rather than which inputs it is enforced on.
 
\begin{table}[h]
\centering
\caption{Constraint surface on WMDP-Cyber (RMU). Retain-side is GRIP as
described in Section~\ref{sec:method} and reproduces the Expert-Specific
row of Table~\ref{fig:ablation}. Unconstrained-baseline reference values
are FA $0.26$, RA $0.35$, RS $0.43$.}
\label{tab:constraint_side}
\small
\begin{tabular}{lcccc}
\toprule
\textbf{Constraint surface} & \textbf{FA}$\downarrow$ & \textbf{RA}$\uparrow$ & \textbf{RS}$\uparrow$ & \textbf{Time}$\downarrow$ \\
\midrule
Forget-side only        & 0.26 & 0.41 & 0.51 & 1.31$\times$ \\
Combined forget+retain  & 0.31 & 0.64 & 0.96 & 2.10$\times$ \\
\textbf{Retain-side (GRIP)} & \textbf{0.24} & \textbf{0.65} & \textbf{0.96} & \textbf{1.71}$\times$ \\
\bottomrule
\end{tabular}
\end{table}

As shown in Table~\ref{tab:constraint_side}, constraining forget-set
routing yields RA of $0.41$, barely above the $0.35$ of the unconstrained
baseline and far below the $0.65$ of retain-side preservation, confirming
that preserving forget-side routing protects little utility. The combined
constraint degrades forget accuracy to $0.31$ at more than double the
training cost, mirroring the over-restriction seen in the ``Full Null Space'' row
of Table~\ref{fig:ablation}. Retain-side preservation therefore achieves the best trade-off
(FA $0.24$, RA $0.65$, RS $0.96$), consistent with the argument in Section~\ref{sec:method} that
because router rows are shared across forget and retain inputs, a
retain-side constraint already limits forget-query rerouting and an
explicit forget-side term adds restriction without adding protection.

\subsection{Per-Layer Threshold Selection}
\label{app:per_layer_eps}

The sweep in Table~\ref{tab:threshold_sensitivity} applies a single
$\epsilon_{\mathrm{null}}$ globally. Because representation drift
accumulates with depth (Section~\ref{sec:method_b}), the spectrum of
$C_{j,\ell}$ that the threshold cuts is not uniform across the network,
so the value that best trades routing stability against plasticity need
not be the same at every layer. We resolve the sweep by depth to test
this.

\paragraph{Protocol.} We partition the $L = 48$ MoE layers into three
depth bands, vary $\epsilon_{\mathrm{null}}$ within one band at a time
while holding the remaining bands at the global default $10^{-2}$, and
report the resulting whole-model metrics. The $10^{-2}$ row therefore
coincides with the global configuration of
Table~\ref{tab:threshold_sensitivity} and serves as the reference. All
runs use WMDP-Cyber with RMU at matched forget accuracy, and are
single-seed, so differences below roughly $0.02$ should not be
over-interpreted.

\begin{table}[h]
\centering
\caption{Per-band threshold sweep on WMDP-Cyber (RMU). Each cell reports
RS\,/\,FA\,/\,RA with $\epsilon_{\mathrm{null}}$ applied to that band
only and all other bands held at $10^{-2}$. The $10^{-2}$ row reproduces
the global configuration of Table~\ref{tab:threshold_sensitivity}.}
\label{tab:per_layer_eps}
\small
\begin{tabular}{lccc}
\toprule
\textbf{$\epsilon_{\mathrm{null}}$} & \textbf{Early (1--16)} & \textbf{Mid (17--32)} & \textbf{Late (33--48)} \\
\midrule
$10^{-4}$ & 0.97\,/\,0.29\,/\,0.62 & 0.96\,/\,0.28\,/\,0.63 & 0.97\,/\,0.27\,/\,0.62 \\
$10^{-3}$ & 0.96\,/\,0.26\,/\,0.62 & 0.96\,/\,0.25\,/\,0.64 & 0.96\,/\,0.25\,/\,0.64 \\
$10^{-2}$ & 0.96\,/\,0.24\,/\,0.65 & 0.96\,/\,0.24\,/\,0.65 & 0.96\,/\,0.24\,/\,0.65 \\
$10^{-1}$ & 0.92\,/\,0.24\,/\,0.59 & 0.92\,/\,0.24\,/\,0.58 & 0.93\,/\,0.24\,/\,0.59 \\
\bottomrule
\end{tabular}
\end{table}

\paragraph{The optimum does not shift with depth.}
As shown in Table~\ref{tab:per_layer_eps}, $\epsilon_{\mathrm{null}} =
10^{-2}$ remains the preferred setting in every band. Tightening buys
marginal routing stability at the cost of erasure,
while loosening to $10^{-1}$ leaves forget accuracy at $0.24$ but costs
both routing stability and roughly six points of retain accuracy.  We therefore find no
evidence that optimal thresholds diverge with depth at this granularity,
which justifies the global setting used throughout the main text.

\paragraph{Layer-adaptive scheduling.}
Because the per-band optima coincide, a schedule built from them would
reduce to the global default, and these results give no reason to prefer
layer-adaptive selection at band granularity. The asymmetry above does
suggest where a finer-grained schedule might pay: late layers absorb
tightening cheaply, so a threshold that decreases with depth is the
direction worth exploring, rather than the increasing one that a naive
drift argument would suggest. Establishing that requires per-layer or
per-expert search rather than the three-way partition used here, and the
bands interact through drift so such a search cannot be decomposed band
by band. We leave this to future work.

\end{document}